\documentclass[journal]{IEEEtran}
\IEEEoverridecommandlockouts

\usepackage[numbers,sort&compress]{natbib}
\usepackage{amsmath,amssymb,amsfonts, bm}
\usepackage[ruled]{algorithm2e}
\usepackage{subfigure}
\usepackage{graphicx}
\usepackage{caption}
\usepackage{textcomp}
\usepackage{xcolor}
\usepackage{authblk}
\usepackage[marginal]{footmisc}
\usepackage{multirow}
\usepackage{multicol}
\usepackage{makecell}
\usepackage{diagbox}
\def\BibTeX{{\rm B\kern-.05em{\sc i\kern-.025em b}\kern-.08em
    T\kern-.1667em\lower.7ex\hbox{E}\kern-.125emX}}

\title{Locality Relationship Constrained Multi-view Clustering Framework}

\author{Xiangzhu Meng\thanks{X. Meng is with Center for Research on Intelligent Perception and Computing, Institute of Automation, Chinese Academy of Sciences, Beijing, China (xiangzhu.meng@cripac.ia.ac.cn).}, Wei Wei\IEEEauthorrefmark{1}\thanks{W. Wei is with Center for Energy, Environment \& Economy Research, Tourism Management School of Zhengzhou University, Zhengzhou, China (weiwei123@zzu.edu.cn).}, and Wenzhe Liu\thanks{W. Liu is with the School of Computer Science and Technology, Dalian University of Technology, Dalian, China (liuwenzhe@mail.dlut.edu.cn).}}

\begin{document}

\maketitle

\begin{abstract}
In most practical applications, it's common to utilize multiple features from different views to represent one object. Among these works, multi-view subspace-based clustering has gained extensive attention from many researchers, which aims to provide clustering solutions to multi-view data. However, most existing methods fail to take full use of the locality geometric structure and similarity relationship among samples under the multi-view scenario. To solve these issues, we propose a novel multi-view learning method with locality relationship constraint to explore the problem of multi-view clustering, called Locality Relationship Constrained Multi-view Clustering Framework (LRC-MCF). LRC-MCF aims to explore the diversity, geometric, consensus and complementary information among different views, by capturing the locality relationship information and the common similarity relationships among multiple views. Moreover, LRC-MCF takes sufficient consideration to weights of different views in finding the common-view locality structure and straightforwardly produce the final clusters. To effectually reduce the redundancy of the learned representations, the low-rank constraint on the common similarity matrix is considered additionally. To solve the minimization problem of LRC-MCF, an Alternating Direction Minimization (ADM) method is provided to iteratively calculate all variables LRC-MCF. Extensive experimental results on seven benchmark multi-view datasets validate the effectiveness of the LRC-MCF method.
\end{abstract}

\begin{IEEEkeywords}
Multi-view clustering, Locality Relationship Constraint, Alternating Direction Minimization
\end{IEEEkeywords}

\section{Introduction}
With the rapid development of the information era, one common object can usually be represented by different viewpoints \cite{zhang2019multi, wang2019gmc}. For example, the document (or paper) can be represented by different languages \cite{Amini2009Learning, bisson2012co}. Generally, multiple views can contain more meaningful and comprehensive information than single view, which provide some knowledge information not involved in single view. The fact can show that, properly using multi-view information is beneficial for improving the practical performance in most situations. However, most traditional methods mainly focus on the single view, which cannot be directly used to deal with multi-view data. Therefore, multi-view learning \cite{li2018survey, zhao2017multi, Xu2015Multi} methods has been proposed in different tasks, such as classifications \cite{kan2016multi, Wang2019A, wang2020kernelized}, clustering \cite{Zheng2018Binary, yang2018multi}, etc. Notably, multi-view clustering \cite{fu2020overview, yang2018multi-cluster, chao2017survey} has become an important research field, which aims at producing satisfied performance by integrating the diversity information among different views. An easy yet effective way is to concatenate all views as one common view, then utilize traditional single view methods to solve the clustering task. However, this manner not only neglects the specific statistical property in each view , but cannot take full use of the consistency and complementary information from multiple views.

To solve above issues, plenty of multi-view clustering methods have been widely proposed in the past ten years. A class of representative multi-view methods \cite{xia2010multiview,Nie2016Param,Nie2017Auto,Huang2018Self, tian19a} investigate to construct one common latent view shared by all views, which can integrate the diversity and complementary information from different views into the common view and then partition it into the ideal groups. For example, Auto-Weighted Multiple Graph Learning (AMGL)\cite{Nie2016Param} method automatically assigns an ideal weight for each view in integrating multi-view information according to the importance of each view. Nevertheless, they cannot guarantee the complementarity information among different views so that rich multi-view information cannot be further fully utilized. Therefore, another class of multi-view clustering methods are proposed to further discover the complementary information among different views. These works based on Canonical Correlation Analysis (CCA) \cite{hardoon2004canonical} and Hilbert-Schmidt Independence Criterion (HSIC) \cite{gretton2005measuring} are two classes of representative multi-view clustering methods. The former \cite{rupnik2010multi, sharma2012generalized, kan2016multi-view} employs CCA to project the pairwise views into the common subspace by maximizing the correlation between two views, and the latter \cite{Niu2014Iterative, Cao2015Diversity, Zhang2016Flexible} explores complementary information based on HSIC term by mapping variables into a reproducing kernel Hilbert space. Apart from these above works, co-training strategy \cite{Wang2010A, kumar2011co2} is also considered as an effective tool to enforce different views to learn from each other. Unfortunately, these methods may produce unsatisfactory results when the gap of different views is varying large. Although such methods have achieved excellent performances in some situations, most of them always focus on either the intrinsic structure information in each view or geometric structures from other views. To discover the geometric structures information in multi-view data, those works \cite{wang2019co, wang2020multi, 9106791, wang2020Discriminative, zhu2020cagnn} have been developed to learn the intrinsic structure in each view. However, the geometric structures information in these works usually depend on artificial definition, which cannot learn the geometric relation among samples from multi-view data. To sum up, multi-view clustering is still an open yet challenging problem.

To solve these issues, we propose a novel multi-view learning method, called Locality Relationship Constrained Multi-view Clustering Framework (LRC-MCF), to explore the problem of multi-view clustering. LRC-MCF simultaneously explores the diversity, geometric, consensus and complementary information among different views, which attempts to capture the locality relationship information in each view, and then fuse the locality structure information in all views into one common similarity relationships among multiple views. In this way, we not only take use of the diversity and geometric information among multiple views, but enforce difference different views to learn with each other by the common similarity relationships. Note that there is different clustering performance for each view, thus LRC-MCF adaptively assign the suitable weights for different views in finding the common-view locality structure and straightforwardly produce the final clusters. To effectually reduce the redundancy of the learned representations, the low-rank constraint on the common similarity matrix is additionally considered. To solve the minimization problem of LRC-MCF, the optimization algorithm based on the alternating direction minimization strategy is provided to iteratively calculate all variables LRC-MCF. Finally, extensive experimental results on seven benchmark multi-view datasets show the effectiveness and superiority of the LRC-MCF method.

\subsection{Contributions}
The major contributions of this paper can be summarized as follows:
\begin{itemize}
    \item LRC-MCF simultaneously explores the diversity, geometric, consensus and complementary information among different views, by capturing the locality relationship information and the common similarity relationships among multiple views.

    \item LRC-MCF takes sufficient consideration to weights of different views in finding the common-view locality structure, and adds the low-rank constraint on the common similarity matrix to effectually reduce the redundancy of the learned representations.

    \item To solve the minimization problem of LRC-MCF, an Alternating Direction Minimization (ADM) method is provided to iteratively calculate all variables LRC-MCF, which can convergence in limited iterations.

    \item The experimental results on seven benchmark datasets demonstrate effectiveness and superiority of LRC-MCF, which outperforms its counterparts and achieves comparable performance.
\end{itemize}

\subsection{Organization}
The rest of the paper is organized as follows: in Section \uppercase\expandafter{\romannumeral2}, we provide briefly some related methods which have attracted extensive attention; in Section \uppercase\expandafter{\romannumeral3}, we describe the construction procedure of LRC-MCF and optimization algorithm for LRC-MCF; in Section \uppercase\expandafter{\romannumeral4}, extensive experiments on text and image datasets demonstrate the effectiveness of our proposed approach; in Section \uppercase\expandafter{\romannumeral5}, the conclusion is put forth finally.

\section{Related works}
In this section, we review two classical multi-view learning methods, including Auto-Weighted Multiple Graph Learning, and Co-regularized Multi-view Spectral Embedding.

\subsection{Auto-Weighted Multiple Graph Learning}
Auto-Weighted Multiple Graph Learning (AMGL)\cite{Nie2016Param} is a multi-view framework based on the standard spectral learning, which can be used for multi-view clustering and semi-supervised classification tasks. Let ${\bm{X}^v}=\left\{ {\bm{x}_1^v,\bm{x}_{_{}^2}^v, \ldots ,\bm{x}_N^v} \right\}$ denote the features set in the $v$th view. AMGL aims to find the common embedding $U \in {\mathbb{R}^{N \times k}}$ as follows:
\begin{equation}\label{amgl}
\begin{split}
&\mathop {\max }\limits_{\bm{U} \in \mathbb{C}} \sum\limits_{v = 1}^M \sqrt{tr({\bm{U}}^{^T}{\bm{L}^v}{\bm{U}})}  \\
\end{split}
\end{equation}
where ${\bm{L}^v}$ denotes the normalized graph Laplacian matrix  in the $v$th view, $trace(\cdot)$ denotes the trace of the matrix, and $k$ is the number of clusters. $\mathbb{C}$ deontes the different constraints in the Eq. (\ref{amgl}), if $\mathbb{C}$ is ${\bm{U}}^T {\bm{U}}=\bm{I}$, the framework is used to solve the clustering tasks; if $\mathbb{C}$ contains the labels information, the framework is also used for the semi-supervised classification task. Applying the Lagrange Multiplier method to solve the Eq. (\ref{amgl}), the adaptive weight mechanism is derived to allocate suitable weights for different views. Therefore, the multi-view framework can be transformed as the following problem:
\begin{equation}
\begin{split}
&\mathop {\max }\limits_{\bm{U} \in \mathbb{C}} \sum\limits_{v = 1}^M {\alpha}^v {tr({\bm{U}}^{^T}{\bm{L}^v}{\bm{U}})}  \\
\end{split}
\end{equation}
where
\begin{equation}
\begin{split}
&  {\alpha}^v = 1/\left(2\sqrt{tr({\bm{U}}^{^T}{\bm{L}^v}{\bm{U}})}\right)\\
\end{split}
\end{equation}
According to the Eq. (\ref{amgl}), if the $v$th view is good, then $tr({\bm{U}}^{^T}{\bm{L}^v}{\bm{U}})$ should be small, and thus the learned weight ${\alpha}^v$ for the $v$th view is large. Correspondingly, a bad view will be also assigned a small weight. Therefore, AMGL optimizes the weights meaningfully and can obtain better result than the classical combination approach which assigns equal weight to all the views.

\subsection{Co-regularized Multi-view Spectral Clustering}
Co-regularized Multi-view Spectral Clustering (Co-reg)\cite{kumar2011co} is a spectral clustering algorithm under multiple views, which achieves this goal by co-regularizing the clustering hypotheses across views. Co-reg aims to propose a spectral clustering framework for multi-view setting. To achieve this goal, Co-reg works with the cross-view assumption that the true underlying clustering should assign corresponding points in each view to the same cluster. For the example of two-view case for the ease of exposition, the cost function for the measure of disagreement between two clusters of the learned embedding $\bm{U}^v$ and $\bm{U}^w$ in the $p$th view and the $q$th view could be defined as follows:
\begin{equation}\label{co_regularizaton1}
    D\left( {{\bm{U}^v},{\bm{U}^w}} \right) =  \left\| \frac{\bm{K}_{\bm{U}^v}}{\left\| \bm{K}_{\bm{U}^v} \right\|_F^2}-\frac{\bm{K}_{\bm{U}^w}}{\left\| \bm{K}_{\bm{U}^w}\right\|_F^2} \right\|_F^2
\end{equation}
where $\bm{K}_{\bm{U}^v}$ and $\bm{K}_{\bm{U}^w}$ denote the similarity matrix for the $p$th view and the $q$th view, respectively. For the convenience of solving the solution, linear kernel is chosen as the similarity measure. Co-reg builds on the standard spectral clustering by appealing to the co-regularized framework, which makes the clustering relationships on different views agree with each other. Therefore, combining Eq. (\ref{co_regularizaton1}) with the spectral clustering objectives of all views, we could get the following joint maximization problem for $M$ views:
\begin{equation}
\begin{split}
&\mathop {\min }\limits_{{\bm{U}^1},{\bm{U}^2}, \ldots ,{\bm{U}^M}} \sum\limits_{v = 1}^M {tr({\bm{U}^v}^{^T}{\bm{L}^v}{\bm{U}^v})} + \lambda \sum\limits_{1 \le v \neq w \le M} {D\left( {{\bm{U}^v},{\bm{V}^w}} \right)} \\
&\hspace{3em}s.t.\hspace{1.5em}{\bm{U}^v}^{^T}{\bm{U}^v}{ =  I,}  1 \leq v \leq M  \\
\end{split}
\end{equation}
where the first term is  the spectral clustering objectives, $\lambda$ is a the non-negative hyperparameter to trade-off the spectral clustering objectives and the spectral embedding disagreement terms across different views. In this way, Co-reg implements a spectral clustering framework for multi-view setting.

\section{The Proposed Approach}
\subsection{Locality relationship preserving}
Given two samples $\bm{x}^v_i$ and $\bm{x}^v_i$ in the $v$th view, the distance between two samples is denoted as $d(\bm{x}^v_i, \bm{x}^v_j)$, which is usually calculated by Mahalanobis distance. Eulidean distance, $L_1$-norm, etc. Intuitively, the distance between samples reflects the relationship between samples, which is of vital importance to construct the local geometric structure for the set of samples. For the $v$th view, conventional similarity relationship between two samples is usually defined as $\bm{S}_{i,j}^v = \exp{(-\frac{d(\bm{x}^v_i, \bm{x}^v_j)}{\sigma})}$, where $\sigma$ is a hyper-parameter. However, one major disadvantage of this manner is that the hyper-parameter is hard to set in practice due to the noise and outliers in the data. Therefore, how to learn the suitable similarity matrix is of vital importance. Meanwhile, the learnt similarity matrix should be subject to such a condition that a smaller distance between two data points corresponds to a large similarity value, and a larger distance between two data points corresponds to a small similarity value. Toward this end, we model the problem as follows:
\begin{equation}\label{eq1}
\begin{split}
&\mathop {\min }\limits_{{\bm{S}^v}} \sum\limits_{i= 1}^N {\sum\limits_{j=1}^N} {d(\bm{x}^v_i, \bm{x}^v_j)\bm{S}_{i,j}^v} + \lambda \left\| \bm{S}^v \right\|_F^2  \\
& s.t. \bm{S}^v_{i,j} \geq 0\\
\end{split}
\end{equation}
where the second term is a regularization term on $\bm{S}^v$, and $\lambda$ is a hyperparameter to balance these two terms. To further analyze $\bm{S}^v$, we add the normalization on $\bm{S}^v$ in the Eq. (\ref{eq1}), i.e. $\bm{1}^T {\bm{s}^v_i}=1$, which makes the second term constant. Then, the above problem can be transformed as follows:
\begin{equation}
\begin{split}
&\mathop {\min }\limits_{{\bm{S}^v}} \sum\limits_{i,j= 1}^N {\sum\limits_{j=1}^N} {d(\bm{x}^v_i, \bm{x}^v_j)\bm{S}_{i,j}^v} + \lambda \left\| \bm{S}^v \right\|_F^2 \\
& s.t. \bm{S}^v_{i,j} \geq 0, \bm{1}^T {\bm{S}^v_i}=1\\
\end{split}
\end{equation}

Inspired by sparse representation method, we attempt to additionally add the sparsity constraint on $\bm{S}^v$, which encourages the similarity matrix $\bm{S}^v$ to have more zero elements. Usually, $L_0$ or $L_1$ norm could be used to implement the sparsity constraint, but this manner might neglect the local structure in representation space. Based on the fact that the sample point could be inflected by its some nearest neighbors, we use the local structure information to construct the sparsity constraint for $\bm{S}^v$, thus we can obtain the following equation based on locality relationship:
\begin{equation}\label{sparsity_constraint}
\begin{split}
&\mathop {\min }\limits_{{\bm{S}^v}} \sum\limits_{i= 1}^N {\sum\limits_{j=1}^N} {d(\bm{x}^v_i, \bm{x}^v_j)\bm{S}_{i,j}^v} + \lambda \left\| \bm{S}^v \right\|_F^2 \\
& s.t. \bm{s}^v_{i,j} \geq 0, \bm{1}^T {\bm{S}^v_i}=1\\
& \hspace{1.5em} \bm{S}^v_{i,j}=0, if \bm{x}_j^v \notin \mathcal{N}({\bm{x}}_i^v)\\
\end{split}
\end{equation}
where $\mathcal{N}(\bm{x}^v_i)$ denotes the $K$ nearest neighbors set of $\bm{x}^v_i$.

\subsection{Multi-view information integration}
For the multi-view setting, a naive way is to incorporate all views directly according to the definition in Eq. (\ref{sparsity_constraint}), which can be expressed as follows:
\begin{equation}
\begin{split}
&\mathop {\min }\limits_{{\bm{S}^v}} \sum_{v=1}^M{ \sum\limits_{i= 1}^N {\sum\limits_{j=1}^N} {d(\bm{x}^v_i, \bm{x}^v_j)\bm{S}_{i,j}^v}} + \lambda \sum_{v=1}^M {\left\| \bm{S}^v \right\|_F^2} \\
& s.t. \forall{v}, \bm{S}^v_{i,j} \geq 0, \bm{1}^T {\bm{S}^v_i}=1\\
& \hspace{1.5em} \bm{S}^v_{i,j}=0, if \bm{x}_j^v \notin \mathcal{N}({\bm{x}}_i^v)\\
\end{split}
\end{equation}
However, this manner is equal to solve the problem for all views separately, which will fail to integrate multi-view features and make favorable use of the complementary information from multiple views. To solve this issue, we firstly make such hypothesis that similarities among the instances in each view and the centroid view should be consistent under the novel representations. This hypothesis means that all similarity matrices from multiple views should be consistent with the similarity of the centroid view, which is implemented by aligning the similarities matrix computed from the centroid view and the $v$th view. Therefore, we utilize the following cost function as a measurement of disagreement between the centroid view and the $v$th view:
\begin{equation}\label{difference}
\begin{split}
D\left( {{\bm{S}^{\ast}},{\bm{S}^v}} \right) = \left\| {{\bm{S}^{\ast}} - {\bm{S}^v}} \right\|_F^2\\
\end{split}
\end{equation}
where $\bm{S}^{\ast}$ denotes the similarity matrix in the centroid view.

To integrate rich information among different features, we could obtain the following optimization problem by adding up cost function in the Eq. (\ref{difference}) from all views:
\begin{equation}
\begin{split}
&\mathop{\min }\limits_{{\bm{S}^{\ast}}} \sum\limits_{v = 1}^M {\left\| {{\bm{S}^{\ast}} - {\bm{S}^v}} \right\|_F^2} \\
& s.t.\hspace{0.5em} \bm{S}^{\ast}_{i,j} \geq 0, \bm{1}^T {\bm{S}^{\ast}_i}=1\\ \\
\end{split}
\end{equation}

In most practical applications, different views cannot play the same role in fusing different similarity matrixes. To reflect the importance of different features in integrating multi-view information, we can obtain the following optimization problem by allocating ideal weights $\bm{w}$ for all views:
\begin{equation}\label{integration}
\begin{split}
&\mathop{\min }\limits_{\bm{w}, \bm{S}^{\ast}} \sum\limits_{v = 1}^M {\bm{w}^v \left\| {{\bm{S}^{\ast}} - {\bm{S}^v}} \right\|_F^2} \\
& s.t.\hspace{0.5em} \bm{S}^{\ast}_{i,j} \geq 0, \bm{1}^T {\bm{S}^{\ast}_i}=1\\
& \hspace{1.5em} \bm{w}^v \geq 0, \bm{1}^T \bm{w}=1 \\
\end{split}
\end{equation}
where $\bm{w}^v$ can reflect the importance of the $v$th view in obtaining the common similarity matrix $\bm{S}^{\ast}$.

\subsection{Multi-view clustering based on Locality relationship preserving}
To simultaneously exploit the diversity and consensus information among different views, we can obtain the multi-view clustering framework by combining the Eq. (\ref{sparsity_constraint}) and Eq. (\ref{integration}):

\begin{equation}
\begin{split}
&\mathop {\min }\limits_{\bm{w}, \bm{S}^{\ast}, \{\bm{S}^v\}} \sum_{v=1}^M{ \sum\limits_{i= 1}^N {\sum\limits_{j=1}^N} {d(\bm{x}^v_i, \bm{x}^v_j)\bm{S}_{i,j}^v}} + \lambda \sum_{v=1}^M {\left\| \bm{S}^v \right\|_F^2} \\ & + \sum\limits_{v = 1}^M {{(\bm{w}^v)}^r \left\| {{\bm{S}^{\ast}} - {\bm{S}^v}} \right\|_F^2} \\
& s.t.\hspace{0.5em} \bm{S}^{\ast}_{i,j} > 0, \bm{1}^T {\bm{S}^{\ast}_i}=1\\
& \hspace{1.5em} \bm{w}^v > 0, \bm{1}^T \bm{w}=1 \\
& \hspace{1.5em} \forall{v}, \bm{s}^v_{i,j} > 0, \bm{1}^T {\bm{s}^v_i}=1\\
& \hspace{1.5em} \bm{S}^v_{i,j}=0, if \bm{x}_j^v \notin \mathcal{N}({\bm{x}}_i^v)\\
\end{split}
\end{equation}

Our paper aims to solve the final clustering problem, i.e., producing the clustering result directly on the unified graph matrix $\bm{S}^{\ast}$ without an additional clustering algorithm or step. So far, the unified graph matrix $\bm{S}^{\ast}$ cannot tackle this problem. Now we give an efficient and yet simple solution to achieve this goal by imposing a rank constraint on the graph Laplacian matrix $\bm{L}^{\ast}$ of the unified matrix $\bm{S}^{\ast}$. The works imply that if $rank(\bm{L}^{\ast})=N-c$, the corresponding $\bm{S}^{\ast}$ is an ideal case based on which the data points are partitioned into $c$ clusters directly. Thus, there is no need to run an additional clustering algorithm on the unified graph matrix $\bm{S}^{\ast}$ to produce the final clusters. Motivated by this theorem, we add a rank constraint to the above problem. Then our final  multi-view clustering model can be formulated as follows:
\begin{equation}\label{final_loss}
\begin{split}
&\mathop {\min }\limits_{\bm{w}, \bm{S}^{\ast}, \{\bm{S}^v\}} \sum_{v=1}^M{ \sum\limits_{i= 1}^N {\sum\limits_{j=1}^N} {d(\bm{x}^v_i, \bm{x}^v_j)\bm{S}_{i,j}^v}} + \lambda \sum_{v=1}^M {\left\| \bm{S}^v \right\|_F^2} \\ & + \sum\limits_{v = 1}^M {\bm{w}^v \left\| {{\bm{S}^{\ast}} - {\bm{S}^v}} \right\|_F^2} \\
& s.t.\hspace{0.5em} \bm{S}^{\ast}_{i,j} > 0, \bm{1}^T {\bm{S}^{\ast}_i}=1\\
& \hspace{1.5em} \bm{w}^v > 0, \bm{1}^T \bm{w}=1 \\
& \hspace{1.5em} \forall{v}, \bm{S}^v_{i,j} > 0, \bm{1}^T {\bm{S}^v_i}=1\\
& \hspace{1.5em} \bm{S}^v_{i,j}=0, if \bm{x}_j^v \notin \mathcal{N}({\bm{x}}_i^v)\\
& \hspace{1.5em} rank(\bm{L}^{\ast})=N-c \\
\end{split}
\end{equation}

\subsection{Optimization process}
Obviously, all the variables in the Eq. (\ref{final_loss}) are coupled together, thus solving the multi-view clustering problem to optimize all variable at once is still challenging. In addition, the constraints are not smooth. To overcome these issues, we propose a novel algorithm based on the Alternating Direction Minimization strategy. Under the assumption that other variables have been obtained, we can calculate $\bm{S}^{\ast}$ via the Augmented Lagrange Multiplier scheme. That is to say, one variable is updated when the other variables are fixed, which inspires us to develop an alternating iterative algorithm to solve problem. The specific updated rules are shown below:

\textbf{Fixing $\bm{w}$ and $\bm{S}^{\ast}$, update $\bm{S}^v$:} When $\bm{w}$ and $\bm{S}^{\ast}$ are given, graph matrix $\bm{S}^v$ in the Eq. (\ref{final_loss}) is independent. Thus, we can update $\bm{S}^v$ one by one, formulated as follows:
\begin{equation}
\begin{split}
&\mathop {\min }\limits_{\bm{S}^v} \sum\limits_{i= 1}^N {\sum\limits_{j=1}^N} {d(\bm{x}^v_i, \bm{x}^v_j)\bm{S}_{i,j}^v} + \lambda \left\| \bm{S}^v \right\|_F^2 \\ & + \bm{w}^v \left\| {{\bm{S}^{\ast}} - {\bm{S}^v}} \right\|_F^2 \\
& s.t.\hspace{0.5em} \bm{S}^v_{i,j} \geq 0, \bm{1}^T {\bm{S}^v_i}=1\\
& \hspace{1.5em} \bm{S}^v_{i,j}=0, if \bm{x}_j^v \notin \mathcal{N}({\bm{x}}_i^v)\\
\end{split}
\end{equation}
It's easy to find that the above equation is independent for different $i$. Thus, we can solve the following problem separately for each $\bm{S}_i^v$:
\begin{equation}
\begin{split}
&\mathop {\min }\limits_{\bm{S}^v} {\sum\limits_{j=1}^N} {d(\bm{x}^v_i, \bm{x}^v_j)\bm{S}_{i,j}^v} + \lambda \left\| \bm{S}^v \right\|_F^2 \\ & + \bm{w}^v \left\| {{\bm{S}_i^{\ast}} - {\bm{S}_i^v}} \right\|_2^2 \\
& s.t.\hspace{0.5em} \bm{S}^v_{i,j} \geq 0, \bm{1}^T {\bm{S}^v_i}=1\\
& \hspace{1.5em} \bm{S}^v_{i,j}=0, if \bm{x}_j^v \notin \mathcal{N}({\bm{x}}_i^v)\\
\end{split}
\end{equation}
By relaxing the constraint, we can transform the above equation as follows:
$\bm{S}_i^v$:
\begin{equation}
\begin{split}
&\mathop {\min }\limits_{\bm{S}^v} {\sum\limits_{j=1}^N} {d(\bm{x}^v_i, \bm{x}^v_j)\bm{S}_{i,j}^v} + \lambda \left\| \bm{S}_i^v \right\|_2^2 \\ & + \bm{w}^v \left\| {{\bm{S}_i^{\ast}} - {\bm{S}_i^v}} \right\|_2^2 \\
& s.t.\hspace{0.5em} \bm{S}^v_{i,j} \geq 0, \bm{1}^T {\bm{S}^v_i}=1\\
\end{split}
\end{equation}
Using Lagrangian multiplier method and KKT condition to solve it, we can obtain the following solution for $\bm{S}_{i,j}^v$:
\begin{equation}
    \begin{split}
     \bm{S}_{i,j}^v = \frac{-d(\bm{x}^v_i, \bm{x}^v_j)+2\bm{w}^v+2\lambda \eta}{2\lambda+2\bm{w}^v}
    \end{split}
\end{equation}
where $\eta$ is Lagrangian coefficient of the constraint $\bm{1}^T {\bm{s}^v_i}=1$. Considering the locality constraint that $\bm{s}^v_{i,j}$ must be equal to 0 if $\bm{x}_j^v \notin \mathcal{N}({\bm{x}}_i^v)$, we can further solve the $\eta$ and $\lambda$ in the above equation. Finally, we can get the following updated rule to solve $\bm{S}^v$:
\begin{equation}\label{solve_S_v}
\begin{split}
\bm{S}_{i,j}^v = \left\{ \begin{array}{l}
\frac{d(\bm{x}^v_i, \bm{x}^v_{\hat{i}})-d(\bm{x}^v_i, \bm{x}^v_j)+2\bm{w}^v)(\bm{S}^{\ast}_{i,j}-\bm{S}^{\ast}_{i,\hat{i}})}{
K(d(\bm{x}^v_i, \bm{x}^v_{\hat{i}})-2\bm{S}^{\ast}_{i,\hat{i}})+\sum_{k=1}^{K}(2\bm{w}^v)\bm{S}^{\ast}_{i,j}-d(\bm{x}^v_i, \bm{x}^v_j))}, \\
\ if \ \bm{x}_j^v \in \mathcal{N}({\bm{x}}_i^v) \\
0, \ otherwise
\end{array} \right.
\end{split}
\end{equation}
where $\hat{i}$ is the index of the $(K+1)$th nearest neighbor of $\bm{x}^i$. In this way, we can sequentially update the similarity matrix in each view by the above equation.

\textbf{Fixing $\bm{w}$ and $\{\bm{S}^v\}$, update $\bm{S}^{\ast}$:} When $\bm{w}$ and $\{\bm{S}^v\}$ are given, the objective function in the Eq. (\ref{final_loss}) on $\bm{S}^{\ast}$ can be reduced as follows:
\begin{equation}
\begin{split}
&\mathop {\min }\limits_{\bm{S}^{\ast}} \sum\limits_{v = 1}^M {{(\bm{w}^v)}^r \left\| {{\bm{S}^{\ast}} - {\bm{S}^v}} \right\|_F^2} \\
& s.t.\hspace{0.5em} \bm{S}^{\ast}_{i,j} > 0, \bm{1}^T {\bm{S}^{\ast}_i}=1, rank(\bm{L}^{\ast})=N-c\\
\end{split}
\end{equation}
It is difficult to solve the above equation because the constraint $rank(\bm{L}^{\ast})=N-c$ is nonlinear. Inspired by the work that $\bm{L}^{\ast}$ is positive semi-definite, the low-rank constraint can be achieved $\sum_{i=1}^c \theta_i=0$, where $\theta_i$ is the $i$th smallest eigenvalue of $\bm{L}^{\ast}$. According to Ky Fan’s theorem \cite{bhatia2013matrix}, we can obtain the following equation by combining it with the above equation:
\begin{equation}
\begin{split}
&\mathop {\min }\limits_{\bm{S}^{\ast}, \bm{F}} \sum\limits_{v = 1}^M {{(\bm{w}^v)}^r \left\| {{\bm{S}^{\ast}} - {\bm{S}^v}} \right\|_F^2} + \beta tr(\bm{F} \bm{L}^{\ast} \bm{F}^T)\\
& s.t.\hspace{0.5em} \bm{S}^{\ast}_{i,j} > 0, \bm{1}^T {\bm{S}^{\ast}_i}=1, \bm{F}\bm{F}^T=\bm{I}\\
\end{split}
\end{equation}
To solve the above equation, we iteratively solve $\bm{S}^{\ast}$ and $\bm{F}$. When $\bm{F}$ is fixed, the above equation on $\bm{S}^{\ast}$ can be transformed as follows:
\begin{equation}
\begin{split}
&\mathop {\min }\limits_{\bm{S}^{\ast}} \sum\limits_{v = 1}^M {{(\bm{w}^v)}^r \left\| {{\bm{S}^{\ast}} - {\bm{S}^v}} \right\|_F^2} + \beta tr(\bm{F} \bm{L}^{\ast} \bm{F}^T)\\
& s.t.\hspace{0.5em} \bm{S}^{\ast}_{i,j} > 0, \bm{1}^T {\bm{S}^{\ast}_i}=1\\
\end{split}
\end{equation}
For solving $\bm{S}^{\ast}$, the above equation can be rewritten as follows:
\begin{equation}
\begin{split}
&\mathop {\min }\limits_{\bm{s}_i^{\ast}} \sum\limits_{v = 1}^M {\sum\limits_{i,j = 1}^N {(\bm{w}^v)}^r {({\bm{S}_{i,j}^{\ast}} - {\bm{S}_{i,j}^v})}^2} + \beta \sum\limits_{i,j=1}^N {{\bm{S}_{i,j}^{\ast}{(\bm{f}_{i}^{\ast} - \bm{f}_{j}^{\ast} )}^2}}\\
& s.t.\hspace{0.5em} \bm{S}^{\ast}_{i,j} > 0, \bm{1}^T {\bm{S}^{\ast}_i}=1\\
\end{split}
\end{equation}
Let $\bm{e}$ denote the Eulidian distance matrix of $\bm{F}$, i.e. $\bm{e}_{i,j}={(\bm{f}_{i}^{\ast} - \bm{f}_{j}^{\ast} )}^2$, it's easy to prove that the above optimal problem is equal to the following equation:
\begin{equation}\label{solve_S_ast}
\begin{split}
&\mathop {\min }\limits_{\bm{s}_i^{\ast}} \sum\limits_{v = 1}^M \sum\limits_{i= 1}^N { \left\| {\bm{S}_{i}^{\ast}} - {\bm{S}_{i}^v}+\frac{\beta}{{(\bm{w}^v)}^r} \bm{e}_i \right\|}_2^2\\
& s.t.\hspace{0.5em} \bm{S}^{\ast}_{i,j} > 0, \bm{1}^T {\bm{S}^{\ast}_i}=1\\
\end{split}
\end{equation}
The Eq. (\ref{solve_S_ast}) is a standard quadratic constraint problem, which can be effectively solved by optimization tools. When $\bm{S}^{\ast}$ is fixed, the above equation on $\bm{F}$ can be simplified as follows:
\begin{equation}\label{solve_F}
\begin{split}
&\mathop {\min }\limits_{\bm{F}} tr(\bm{F} \bm{L}^{\ast} \bm{F}^T)\\
& s.t. \bm{F}\bm{F}^T=\bm{I}\\
\end{split}
\end{equation}
Obviously, $\bm{L}^{\ast}$ is a symmetric positive-definite matrix. Based on the Ky-Fan theory, $\bm{F}$ in Eq.(\ref{solve_F}) has a global optimal solution, which is given as the eigenvectors associated with the $c$ smallest eigenvalues of $\bm{L}^{\ast}$.

\textbf{Fixing $\{\bm{S}^v\}$ and $\bm{S}^{\ast}$, update $\bm{w}$:} When $\{\bm{S}^v\}$ and $\bm{S}^{\ast}$ are given, the objective function on $\bm{w}$ can be reduced as follows:
\begin{equation}
\begin{split}
&\mathop {\min }\limits_{\bm{w}} \sum\limits_{v = 1}^M {\bm{w}^v \left\| {{\bm{S}^{\ast}} - {\bm{S}^v}} \right\|_F^2} \\
& s.t.\hspace{0.5em} \bm{w}^v > 0, \bm{1}^T \bm{w}=1 \\
\end{split}
\end{equation}
The solution to $\bm{w}$ in the above equation is $\bm{w}^v=1$ corresponding to the $min(D\left( {\bm{S}^{\ast},\bm{S}^v} \right))$ over different views, and $\bm{w}^v=0$ otherwise. This solution means that only one view is finally selected by this method. Therefore, the performance of this method is equivalent to the one from the best view. This solution does not meet our objective on exploring the complementary property of multiple views to get a better embedding than based on a single view. To avoid this phenomenon, we adopt a simple yet effective trick, i.e. we set $\bm{w}^v \leftarrow {(\bm{w}^v)}^r$ with $r \geq 1$. In this way, the novel objective function can be transformed as follows:
\begin{equation}
\begin{split}
&\mathop{\min }\limits_{\bm{w}} \sum\limits_{v = 1}^M {{(\bm{w}^v)}^r \left\| {{\bm{S}^{\ast}} - {\bm{S}^v}} \right\|_F^2} \\
& s.t. \hspace{1.5em} \bm{w}^v \geq 0, \bm{1}^T \bm{w}=1 \\
\end{split}
\end{equation}

By using the Lagrange multiplier $\gamma$ to take the constraint $\bm{1}^T \bm{w}=1$ into consideration, we get the following Lagrange function:
\begin{equation}
\begin{split}
& \mathcal{L}(\bm{w}, \gamma) = \sum\limits_{v = 1}^M {{(\bm{w}^v)}^r \left\| {{\bm{S}^{\ast}} - {\bm{S}^v}} \right\|_F^2} - \gamma (\sum\limits_{v = 1}^M {\bm{w}^v} - 1)
\end{split}
\end{equation}
By setting the partial derivatives of $\mathcal{L}(\bm{w}, \gamma)$ with respect to $\bm{w}$ and $\gamma$ to zeros, $\bm{w}$ can be calculated as follows:
\begin{equation}\label{solve_w}
\begin{split}
 \bm{w}^v = \frac{{( \left\| {{\bm{S}^{\ast}} - {\bm{S}^v}} \right\|_F^2)}^{1/(1-r)}}{\sum_{i=1}^{M}{{ (\left\| {{\bm{S}^{\ast}} - {\bm{S}^v}} \right\|_F^2)}^{1/(1-r)}}}
\end{split}
\end{equation}
Obviously, $\left\| {{\bm{S}^{\ast}} - {\bm{S}^v}} \right\|_F^2$ is a non-negative scala, thus we have $\bm{w}^v \geq 0$ naturally. According to the above equation, we have the following understanding for $r$ in controlling $\bm{w}$. When $r \to \infty$, different $\bm{w}^v$ will be close to each other. When $r \to \infty$, only $\bm{w}^v=1$ corresponding to the minimum value of $D\left( {\bm{S}^{\ast}},{\bm{S}^v}\right)$ over different views, and $\bm{w}^v=0$ otherwise. Therefore, the selection of $r$ should be based on the complementary property of all views.

\begin{algorithm}
\caption{The optimization procedure for LRC-MCF}
\label{algo}
\LinesNumbered
\KwIn{Multi-view features $\{\bm{X}^v,\forall 1\le v \le M \}$, the hyper-parameters $\alpha$ and $\beta$, the distance metric function $d(\cdot, \cdot)$, the number of clustering $c$.}

\For{v=1:M}{
    Initialize $d(\bm{x}^v_i, \bm{x}^v_j)$ according to the distance metric function $d(\cdot, \cdot)$.

    Initialize $\bm{S}^v$ by solving the Eq. (\ref{sparsity_constraint}).

    Initialize $\bm{w}^v$ = 1/M.
}

Fixing $\bm{w}^v$ and $\bm{S}^v$, initialize $\bm{S}^{\ast}$ by solving the Eq. (\ref{solve_S_ast}).

\While{all variables not converged}{
\For{v=1:M}{
    Fixing $\bm{w}^v$ and $\bm{S}^{\ast}$, update $\bm{S}^v$ by the Eq. (\ref{solve_S_v}).

    Fixing $\bm{S}^v$ and $\bm{S}^{\ast}$, update $\bm{w}^v$ by the Eq. (\ref{solve_w}).
}

Fixing $\bm{w}^v$ and $\bm{S}^v$, update $\bm{S}^{\ast}$ by solving the Eq. (\ref{solve_S_ast}).

Fixing $\bm{S}^{\ast}$, update $\bm{F}$ by solving the Eq. (\ref{solve_F}).
}

\KwOut{The common similarity relationship matrix $\bm{S}^{\ast}$.}
\end{algorithm}

To sum up, we can iteratively solve the all variables in Eq. (\ref{final_loss}) according to the aforementioned descriptions, which obtain a local optimal solution of LRC-MCF. For the convenient of readers, we summarize the whole optimization process in the Algorithm {\ref{algo}}. Since problem is not a joint convex problem of all variables, obtaining a globally optimal solution is still an open problem. We solve problem using an alternating algorithm (as Algorithm {\ref{algo}}). As each sub-problem is convex and we find the optimal solution of each sub-problem, thus the algorithm converges because the objective function reduces with the increasing of the iteration numbers. In particular, with fixed other variables, one optimal variable can reduce the value of the objective function. Moreover, we will empirically validate the convergence of our algorithm in the following experiments.

\subsection{Computational Complexity Analysis}
To clearly show the efficiency of LRC-MCF, we provide its computational complexity analysis in this section. Obviously, the computational complexity to solve the all variables in Eq. (\ref{final_loss}) is mainly  dependent on the following five parts: the computational complexity to initialize all variables in Eq. (\ref{final_loss}) is $O(MN^2d+MNK)$, where $d$ is the computational complexity to the distance metric function $d(\cdot, \cdot)$; the computational complexity to update $\bm{S}^v$ by the Eq. (\ref{solve_S_v}) is $O(MNK)$; the computational complexity to update $\bm{S}^{\ast}$ by solving the Eq. (\ref{solve_S_ast}) is $O(CN)$; the computational complexity to update $F$ is $O(CN^2)$; the computational complexity to update $\bm{w}^v$ by the Eq. (\ref{solve_w}) is $O(MN^2)$. Accordingly, the overall computational complexity of Algorithm {\ref{algo}} is about $O(MN^2d+MNK+T(MNK+CN^2+CN+MN^2))$, where $T$ is the iteration times of the alternating optimization procedure in the Algorithm {\ref{algo}}.

\section{Experiments}
To evaluate the effectiveness and superiority of the proposed LRC-MCF, we conduct comprehensive clustering experiments on seven benchmark multi-view datasets in this section. Firstly, we introduce the details of the experiments, including benchmark datasets, compared methods, experiment setting, and evaluation metrics. Then, we perform  all methods on different multi-view datasets to evaluate the performance of LRC-MCF and analysis the experimental results. Finally, we empirically demonstrate the sensitivity analysis and the parameter analysis of LRC-MCF. %

\subsection{Datasets}
To comprehensively demonstrate the effectiveness of the proposed LRC-MCF, we perform all multi-view clustering methods on seven benchmark multi-view datasets, including BBC\footnote[1]{http://mlg.ucd.ie/datasets/segment.html}, BBCSport\footnote[2]{http://mlg.ucd.ie/datasets/segment.html}, 3Source\footnote[3]{http://mlg.ucd.ie/datasets/3sources.html}, Cora\footnote[4]{3http://lig-membres.imag.fr/grimal/data.html}, Handwritten\footnote[5]{https://archive.ics.uci.edu/ml/datasets/One-hundred+palnt+pecies+leaves+data \ +set}, HW2sources\footnote[6]{https://cs.nyu.edu/roweis/data.html}, and NGs\footnote[7]{http://lig-membres.imag.fr/grimal/data.html}. \textbf{BBC} is a documents dataset consisting of 685 documents, where each document is received from BBC news corpora, corresponding to the sports news in five topical areas, and when we treat each topical area as one view, \textbf{BBC} can be regarded as a dataset with five views; \textbf{BBCSport} consists of 544 documents received from the BBC Sport website, where each document is corresponding to the sports news in five topical areas, and \textbf{BBCSport} is also can be regarded as a dataset with five views; \textbf{3Source} contains 169 news, collected from three well-known news organizations (including BBC, Reuters, and Guardian), where each news is manually annotated with one of six labels; \textbf{Cora} contains 2708 scientific publications of seven categories, where each publication document can be described by content and citation, thus \textbf{Cora} could be considered as a two-view benchmark dataset; \textbf{Handwritten} is a hand-written dataset consisting of 2000 samples, where each sample is represented by six different features; \textbf{HW2sources} is a benchmark dataset from the UCI repository, including 2000 samples, which is collected two sources including Mnist hand-written digits and USPS hand-written digits; \textbf{NGs} is a dataset consisting of 500 documents, where each document is pre-proposed by three different methods and manually annotated with one of five labels. For convenience, we summarize the statistic information on these datasets in Table \ref{datasets}.

\begin{table}[htbp]
\caption{The summary information of benchmark datasets}
\label{datasets}
\centering
\renewcommand\arraystretch{1.5}
\begin{tabular*}{0.45\textwidth}{@{\extracolsep{\fill}}lccc}  
\Xhline{1.2pt}
\hline
Datasets &Samples &Classes &Views\\
\hline  
BBC & 685 & 5 & 4\\
BBCSport & 544 & 5 & 2\\
3Source & 169 & 6 & 3\\
Cora & 2708 & 7 & 2\\
Handwritten & 2000 & 10 & 6\\
HW2sources & 2000 & 10 & 2\\
NGs & 500 & 5 & 3\\
\hline
\Xhline{1.2pt}
\end{tabular*}
\end{table}

\subsection{Compared Methods}

We compare the proposed LRC-MCF with the following baseline methods on the above seven benchmark datasets, including two single-view learning methods, one feature concatenation method, and eight multi-view learning methods. The detail information of these comparing methods can be summarized as follows:
\begin{itemize}
    \item \textbf{Single-view learning methods}: Spectral Clustering (SC) \cite{Ulrike2007Spectral} first computes the affinity between each pair of points to construct the similarity matrix, and then make use of the spectrum (eigenvalues) of the similarity matrix to perform dimensionality reduction before clustering in fewer dimensions; Graph regularized Non-negative Matrix Factorizatio (GNMF) \cite{Cai2011Graph} aims to yield a natural parts-based representation for the data and jointly discover the intrinsic geometrical and discriminating structure of the data space.

    \item \textbf{Feature concatenation method}: We first concatenates the features of all views and then employs the traditional clustering method to obtain the clustering results.

    \item \textbf{Multi-view learning methods}: Canonical Correlation Analysis (CCA) \cite{hardoon2004canonical} is utilized to deal with multi-view problems by maximizing the cross correlation between each pair-views; Multi-view Spectral Embedding (MSE) \cite{xia2010multiview} is a multi-view spectral method, which is based on global coordinate alignment to obtain the common representations; Multi-view Non-negative Matrix Factorization (MultiNMF) is proposed in work \cite{liu2013multi} to jointly learn non-negative potential representations from multi-view information and obtain the clustering results by the latent common representations; Co-regularized Multi-view Spectral Clustering (Co-reg) is a multi-view spectral clustering method proposed in work \cite{kumar2011co}, which enforces different view to mutually learn by regularizing different views to be close to each other; Auto-weighted multiple graph learning (AMGL) \cite{Nie2016Param} method, which automatically assigns an ideal weight for each view in integrating multi-view information; Multi-view Dimensionality co-Reduction (MDcR) \cite{2016Flexible} adopts the kernel matching constraint based on Hilbert-Schmidt Independence Criterion to enhance the correlations and penalizes the disagreement of different views; AASC \cite{huang2012affinity} is proposed by aggregating affinity matrices, which attempt to make spectral clustering more robust by relieving the influence of inefficient affinities and unrelated features; WMSC \cite{zong2018weighted} takes advantage of the spectral perturbation characteristics of spectral clustering, uses the maximum gauge angle to estimate the difference between the impacts of distinct spectral clustering, and finally transforms the weight solution problem into a standard secondary planning problem; AWP \cite{nie2018multiview} is an extension of spectral rotation for multiview data with Procrustes Average, which takes the clustering capacity differences of different views into consideration..

\end{itemize}

\subsection{Experiment Setting and Evaluation Metrics}

To implement the clustering of the above datasets, we perform all comparing methods and the proposed LRC-MCF on seven multi-view datasets. Specifically, for single-view learning methods, we choose the best ones from the clustering results of different views, i.e. $\rm{SC}_{best}$ and $\rm{GNMF}_{best}$; for feature concatenation method, we utilize standard Spectral Clustering method \cite{Ulrike2007Spectral} to deal with the concatenated features; for multi-view learning methods, we perform all multi-view methods for given datasets to obtain the corresponding representations and then utilize $K$-means clustering \cite{Hartigan1979} to cluster all samples; for the proposed LRC-MCF, we directly perform the clustering process on the common similarity relationship matrix $\bm{S}^{\ast}$ solved by the Algorithm {\ref{algo}}. Since the clustering \cite{Hartigan1979} algorithm is known to be sensitive to initialization, We run all methods repeatedly 10 times and then obtain the mean value of all indicators for all methods. It's worthy noting that all experiments are performed in the same environment.

Given the label information of all samples, three validation metrics are adapted to validate the performance of clustering method, including accuracy (ACC), normalized mutual information (NMI), and purity (PUR). Specifically, ACC is defined as follows:
\begin{equation}
\begin{split}
&\rm{ACC} = \frac{\sum_{i=1}^N \delta(\bm{y}^i, map(\hat{\bm{y}}_i))}{N} \\
\end{split}
\end{equation}
where $\bm{y}^i$ and $\hat{\bm{y}}_i$ are the truth label and cluster label, respectively. $map(\cdot)$ is the permutation function that maps cluster label to truth label, and $\delta(\cdot, \cdot)$ is the function that equals one if two inputs are equal, and zero otherwise.
Meanwhile, NMI is formulated as follows:
\begin{equation}
\begin{split}
&\rm{NMI} = \frac{\sum_{i=1}^{K} \sum_{j=1}^{K} N_{i, j} log \frac{N_{i, j}}{N_i \hat{N}_j}}{\sqrt{ \sum_{i=1}^{K} N_i log \frac{N_i}{N} \sum_{j=1}^{K} N_j log \frac{\hat{N}_j}{N} }}
\end{split}
\end{equation}
where $N_i$ is the number of samples in the $i$th cluster, $\hat{N}_j$ is the number of samples in the $j$th class, and $N_{i,j}$ is the number of samples in the intersection between the $i$th cluster and the $j$th class.
Finally, PUR is expressed as follows:
\begin{equation}
\begin{split}
&\rm{PUR} = \frac{1}{N} \sum_{i=1}^{K} \max_{1 \le j \le K} \left| \bm{A}_i \cap \hat{\bm{A}}_j \right|\\
\end{split}
\end{equation}
where $\bm{A}_i$ and $\hat{\bm{A}}_j$ are two sets with responding to the $i$th class and the $j$th cluster, and $\left| \cdot \right|$ is the length of set.

Note that the higher values indicate better clustering performance for all evaluation metrics. Different metrics can reflect different properties in the multi-view clustering task, thus we summarize all results on these measures to obtain a more comprehensive evaluation.

\begin{table*}[htbp]
\caption{ACC(\%) results on seven benchmark datasets for all methods}
\label{ACC}
\centering
\renewcommand\arraystretch{1.5}
\begin{tabular*}{0.8\textwidth}{@{\extracolsep{\fill}}lcccccccc}  
\hline
\diagbox{Methods}{Datasets} & BBC & BBCSport & 3Source & Cora & Handwritten & HW2sources & NGs & MEAN\\
\hline  
$\rm{SC}_{best}$ & 43.94 & 50.74 & 39.64 & 35.82 & 92.82  & 58.50 & 28.96 & 50.06 \\
$\rm{GNMF}_{best}$ & 32.12 & 32.12 & 45.56 & 31.91 & 73.80 & 51.10 & 23.01 & 41.37 \\
$\rm{SC}_{concat}$ & 47.26 & 67.22 & 36.45 & 32.14 & 60.03 & 76.24 & 35.98  & 50.76\\
CCA & 48.12 & 38.60 & 41.01 & 44.15 & 73.63 & 74.51 & 27.62 & 49.66\\
MSE & 39.26 & 54.01 & 44.85 & 31.69 & 79.84 & 73.27 & 27.55 & 50.07 \\
MultiNMF & 43.36 & 50.00 & 52.66 & 35.60 & 88.00 & 76.75 & 33.61 & 54.28 \\
Co-reg & 40.73 & 61.58 & 45.56 & 39.11 & 91.00 & 82.15 & 27.45 & 55.37 \\
AMGL & 39.21 & 55.22 & 42.31 & 32.05 & 83.34 & 90.38 & 30.83 & 53.33 \\
MDcR & 81.05 & 80.24 & 70.53 & 54.51 & 61.26 & 86.56 & 59.06 & 70.45  \\
AASC & 38.39 & 62.32 & 36.69 & 33.09 & 84.65 & 83.30 & 71.00 & 58.49 \\
WMSC & 42.92 & 61.21 & 42.60 & 42.25 & 84.50 & 83.40 & 43.00 & 57.12 \\
AWP & 37.37 & 44.85 & 51.48 & 32.39 & \textbf{96.70} & 93.85 & 23.80 & 54.34 \\
LRC-MCF & \textbf{87.45} & \textbf{80.70} & \textbf{78.70} & \textbf{51.18}& 88.25 & \textbf{99.45} & \textbf{98.20}  & \textbf{83.42} \\
\hline
\end{tabular*}
\end{table*}

\begin{table*}[htbp]
\caption{NMI(\%) results on seven benchmark datasets for all methods}
\label{NMI}
\centering
\renewcommand\arraystretch{1.5}
\begin{tabular*}{0.8\textwidth}{@{\extracolsep{\fill}}lcccccccc}  
\hline
\diagbox{Methods}{Datasets} & BBC & BBCSport & 3Source & Cora & Handwritten & HW2sources & NGs & MEAN\\
\hline  
$\rm{SC}_{best}$ & 23.04 & 21.20 & 33.53 & 16.35 & 87.78 & 59.41 & 7.39  & 35.53 \\
$\rm{GNMF}_{best}$ & 21.43 & 19.99 & 29.39 & 19.14 & 74.37 & 57.49 & 8.85  & 30.09 \\
$\rm{SC}_{concat}$ & 35.45 & 52.10 & 35.72 & 19.22 & 61.14 & 80.57 & 22.75 & 43.85 \\
CCA & 30.93 & 25.30 & 35.64 & 26.74 & 68.15 & 60.83 & 13.35 &  37.28 \\
MSE & 26.93 & 34.71 & 46.61 & 18.58 & 80.13 & 72.17 & 13.58 &  41.82 \\
MultiNMF & 27.57 & 44.69 & 43.98 & 26.28 & 80.25 & 68.69 & 20.37 &  44.55 \\
Co-reg & 23.36 & 41.22 & 44.77 & 22.50 & 88.33  & 85.61 & 7.66 & 46.21 \\
AMGL & 26.90 & 39.92 & 45.67 & 18.77 & 81.08 & 88.27 & 19.05 & 45.67\\
MDcR & 61.47 & 63.94 & 64.43 & 31.30 & 66.65 & 77.71 & 50.14 & 59.34  \\
AASC & 20.14 & 39.11 & 30.21 & 16.63 & 87.48 & 86.14 & 48.02 & 46.81 \\
WMSC & 22.10 & 47.12 & 42.00 & 27.96 & 87.44 & 86.50 & 24.84 & 48.28 \\
AWP & 20.06 & 23.15 & 41.53 & 12.26 & \textbf{92.08} & 91.00 & 3.88 & 40.57 \\
LRC-MCF & \textbf{73.62} & \textbf{76.00} & \textbf{70.50} & \textbf{37.85} & \textbf{92.03} & \textbf{98.63} & \textbf{93.92} &  \textbf{77.50}\\
\hline\end{tabular*}
\end{table*}

\begin{table*}[htbp]
\caption{PUR(\%) results on seven benchmark datasets for all methods}
\label{PUR}
\centering
\renewcommand\arraystretch{1.5}
\begin{tabular*}{0.8\textwidth}{@{\extracolsep{\fill}}lcccccccc}  
\hline
\diagbox{Methods}{Datasets} & BBC & BBCSport & 3Source & Cora & Handwritten & HW2sources & NGs  & MEAN\\
\hline  
$\rm{SC}_{best}$ & 49.64 & 55.70 & 56.80 & 40.73 & 93.29 & 62.75 & 29.44 & 55.48 \\
$\rm{GNMF}_{best}$ & 33.43 & 32.99 & 56.80 & 38.40 & 75.10 & 59.10 & 24.88 & 45.81 \\
$\rm{SC}_{concat}$ & 51.42 & 68.66 & 55.38 & 38.43 & 61.62 & 79.33 & 38.86 & 56.24 \\
CCA & 52.63 & 46.47 & 54.97 & 50.37 & 73.69 & 74.51 & 29.33 & 54.57 \\
MSE & 45.40 & 56.58 & 65.56 & 37.98 & 80.71 & 76.96 & 28.70 & 55.98 \\
MultiNMF & 43.65 & 63.42 & 60.36 & 49.74 & 88.00 & 76.75 & 35.99 & 59.70 \\
Co-reg & 49.20 & 66.91 & 62.72 & 45.27 & 91.00 & 86.70 & 28.68 & 61.50 \\
AMGL & 45.40 & 58.93 & 64.97 & 38.30 & 83.34 & 91.22 & 32.96 & 59.30 \\
MDcR & 81.05 & 80.24 & 76.75 & 56.19 & 67.22 & 86.56 & 69.30 & 73.90  \\
AASC & 46.42 & 67.10 & 55.03 & 40.99 & 87.10 & 85.85 & 71.00 & 64.78 \\
WMSC & 48.61 & 68.01 & 60.36 & 47.16 & 87.10 & 86.10 & 46.00 & 63.33 \\
AWP & 45.69 & 51.10 & 63.31 & 35.60 & \textbf{96.70} & 93.85 & 24.40 & 58.66 \\
LRC-MCF & \textbf{87.45} & \textbf{84.38} & \textbf{82.25} & \textbf{52.22} & \textbf{88.05} & \textbf{99.45} & \textbf{98.20} & \textbf{84.57} \\
\hline
\end{tabular*}
\end{table*}

\subsection{Performance Evaluation}
In this section, to show the effectiveness of LRC-MCF, extensive clustering experiments have been conducted on seven multi-view datasets (BBC, BBCSport, 3Source, Cora, Handwritten, and NGs), and the experimental results are summarized in Table \ref{ACC}, Table \ref{NMI} and Table \ref{PUR}, where the bold values represent the superior performance in each table. To comprehensively show the performance of clustering methods, we additionally add up one new column of MEAN in each table, which averages the evaluation indexes of all benchmark datasets for each clustering method.

For BBC and BBCSports datasets, the experimental results show that LRC-MCF can performs better than comparing single-view and multi-view clustering methods in most situations, especially in terms of ACC and PUR. This validates the effectiveness of LRC-MCF in the domain of sports news clustering, and shows that LRC-MCF can capture the locality relationship information and the common similarity relationships among multiple views. Besides, multi-view methods cannot always obtain better performance than single-view methods, which implies that the diversity and complementary information cannot be readily exploited.

For 3Source and Cora datasets, the experimental results show that LRC-MCF can obtain the best performance in terms of ACC, NMI, and PUR. Besides, most multi-view methods outperform single-view methods, which might imply that the diversity information among multiple views is benefit for the improvement of clustering performance. However, these methods cannot get better than RC-MCF, which might be caused by the issue that the geometric and complementary information cannot be made full use of.

For Handwritten and HW2sources datasets, the experimental results also show that LRC-MCF can performs better than comparing single-view and multi-view clustering methods in most situations, which can validate the superiority of LRC-MCF in the domain of image clustering. It implies that the locality relationship information in image clustering tasks is the important factor. Thus, how to exploit the common locality information among multiple views is of great significance.

For NGs dataset, the experimental results show that LRC-MCF can obtain the best performance in terms of ACC, NMI, and PUR. This further validates the effectiveness of LRC-MCF, which shows that LRC-MCF can fully explore the diversity, geometric, consensus and complementary information among different views and take sufficient consideration to weights of different views in finding the common-view locality structure.

Finally, the columns of MEAN in Tables \ref{ACC} - \ref{PUR} validate the comprehensive performance of LRC-MCF, the experimental results show that LRC-MCF can performs better than comparing clustering methods in terms of ACC, NMI, and PUR. It can imply that LRC-MCF not only takes full of the locality geometric structure and similarity relationship among samples under the multi-view scenario, but explores the diversity, consensus and complementary information among different views. Thus, LRC-MCF can obtain the best performance on seven benchmark datasets in most situations.

\subsection{Parameter Analysis}
\begin{figure*}[htbp]
\subfigure[ACC in BBC dataset]{
\centering
\includegraphics[width=0.31\textwidth]{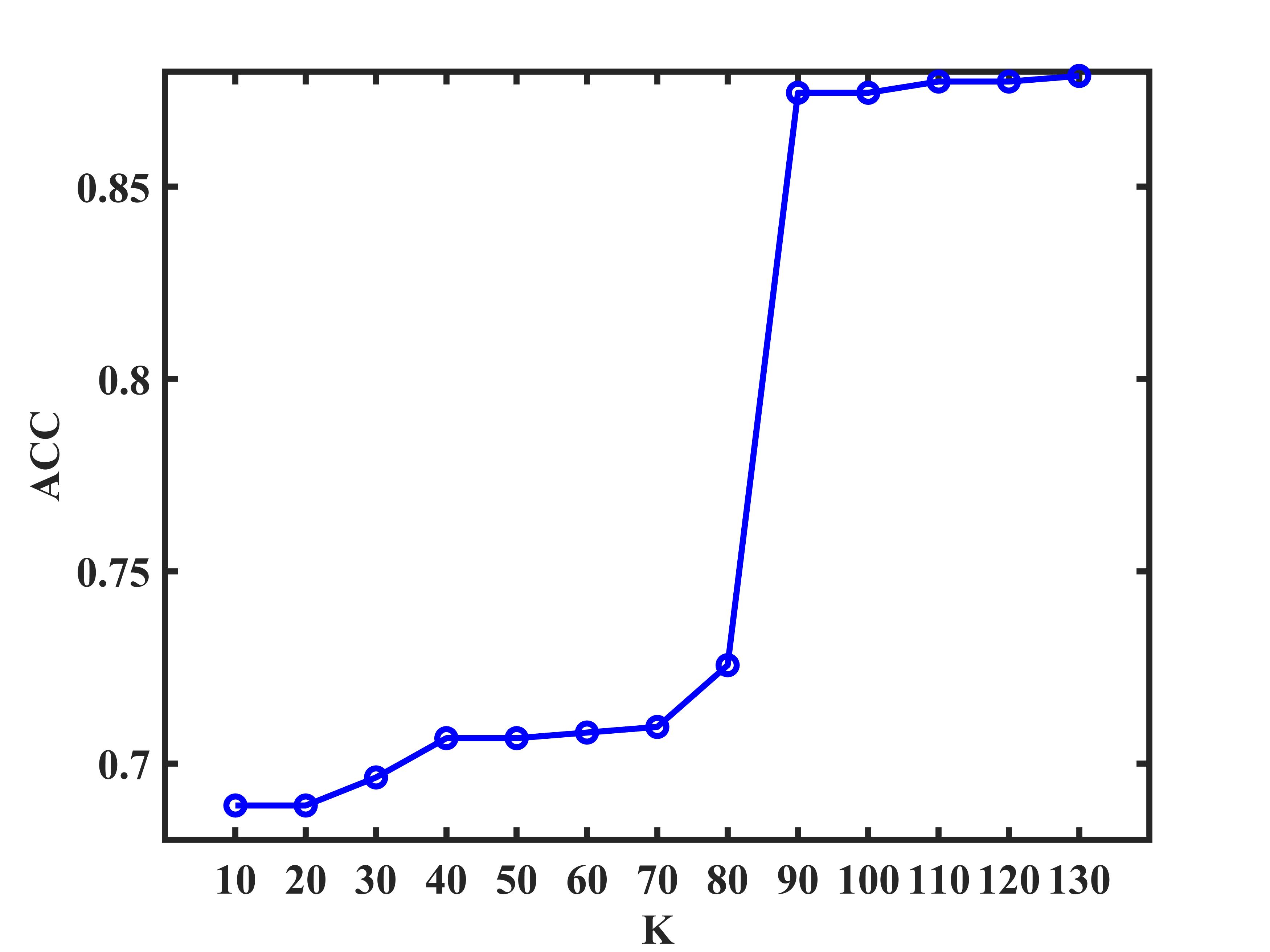}
}
\subfigure[NMI in BBC dataset]{
\centering
\includegraphics[width=0.31\textwidth]{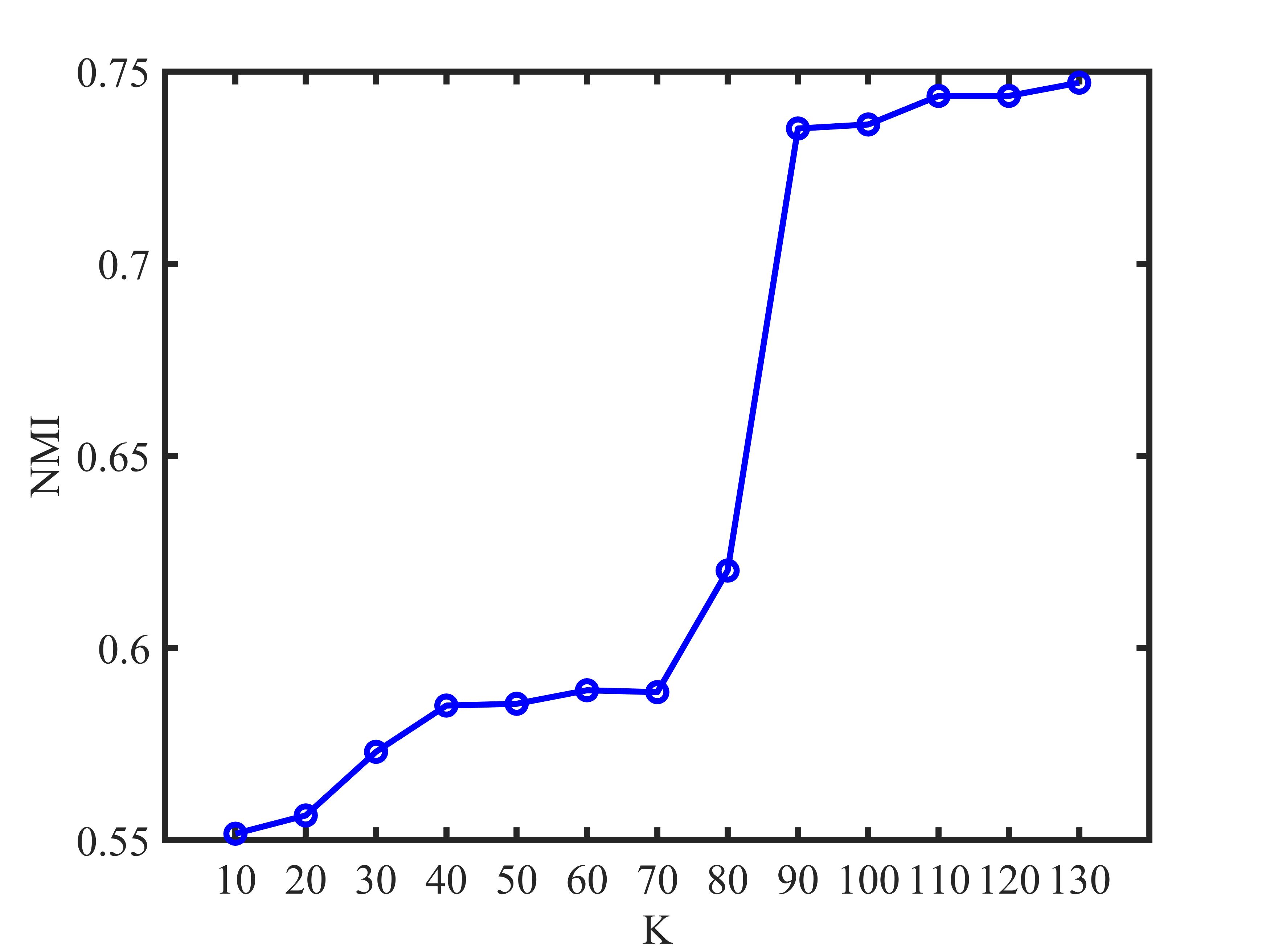}
}
\subfigure[PUR in BBC dataset]{
\centering
\includegraphics[width=0.31\textwidth]{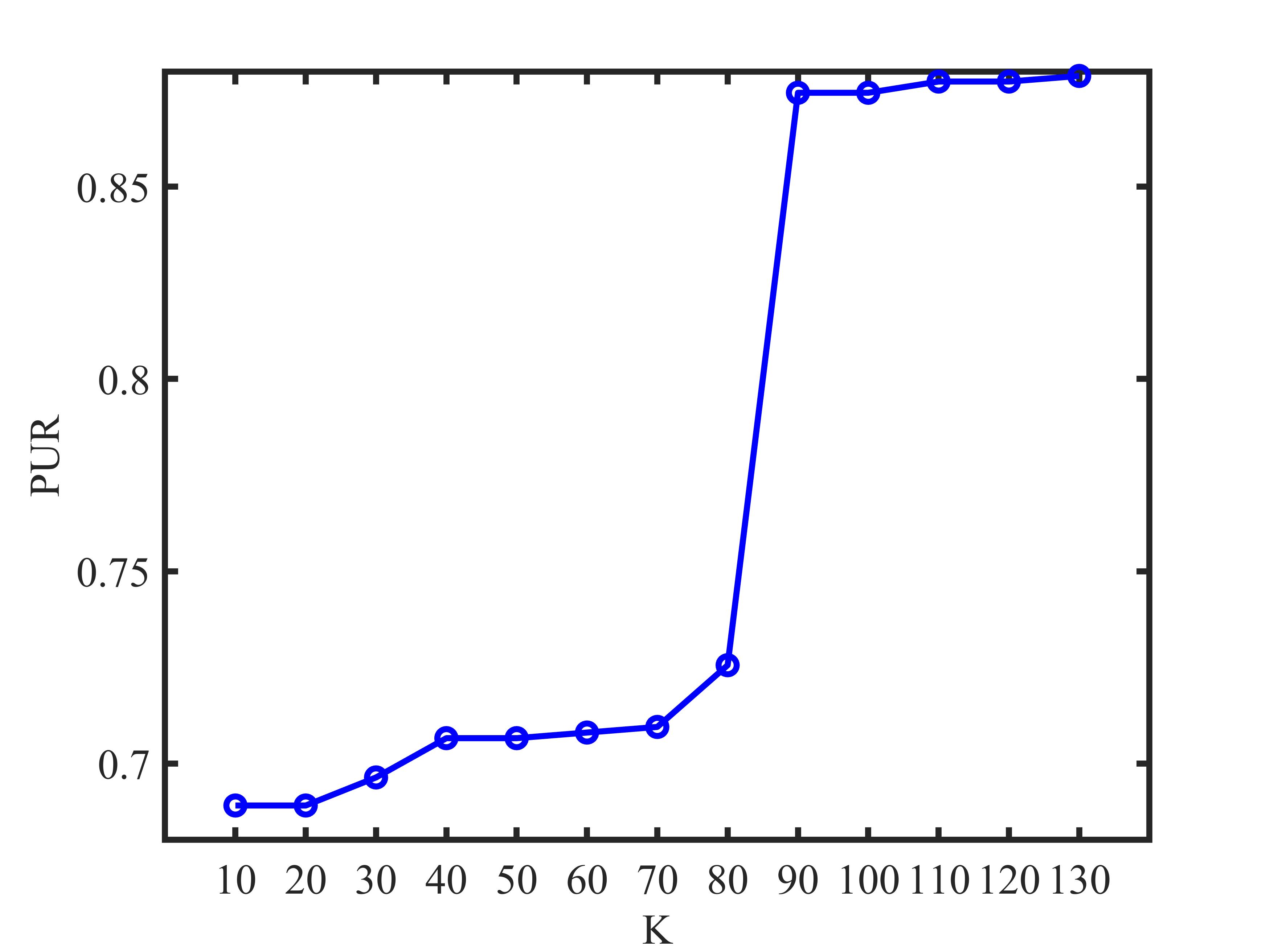}
}

\subfigure[ACC in 3Source dataset]{
\centering
\includegraphics[width=0.31\textwidth]{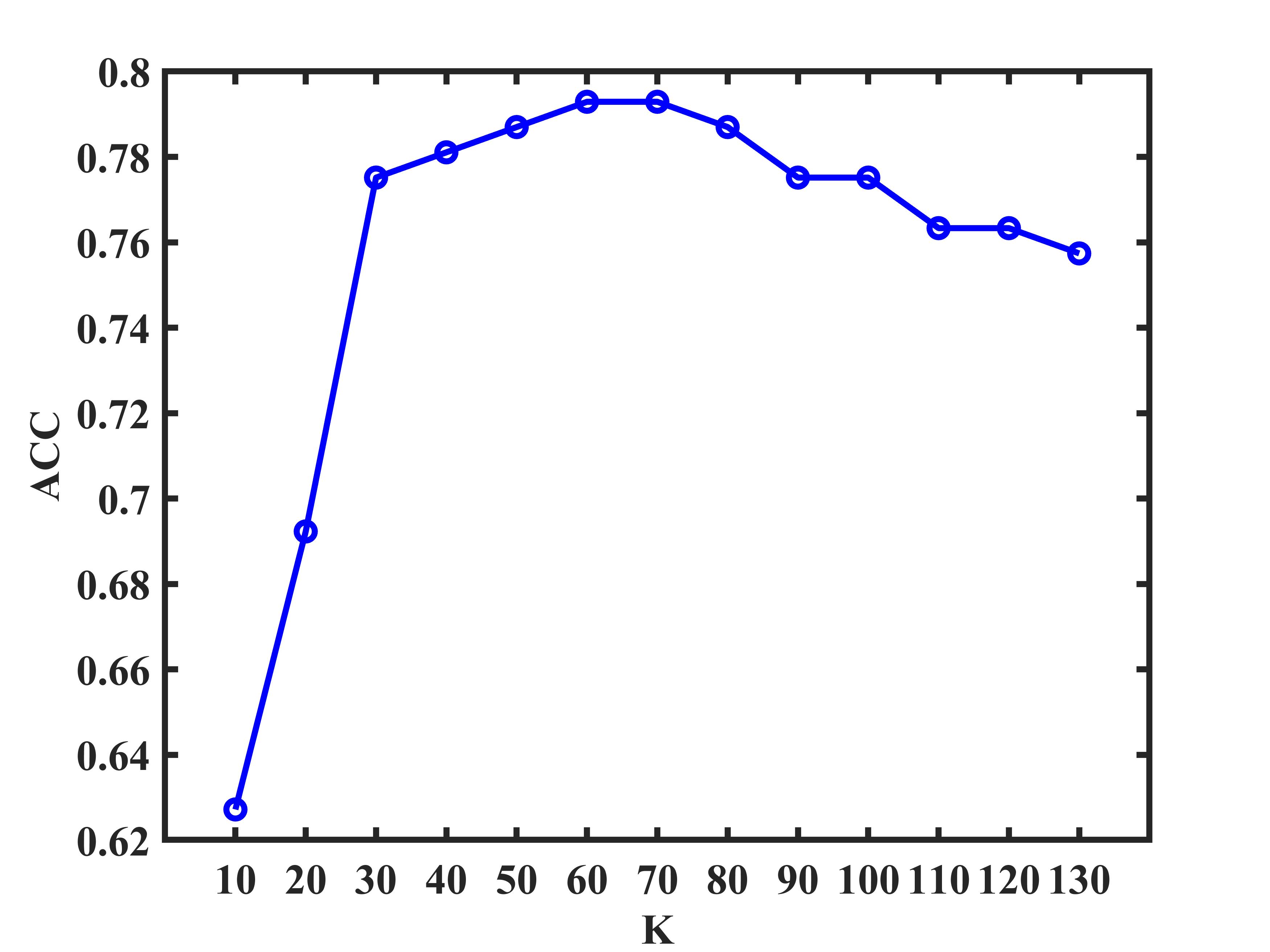}
}
\subfigure[NMI in 3Source dataset]{
\centering
\includegraphics[width=0.31\textwidth]{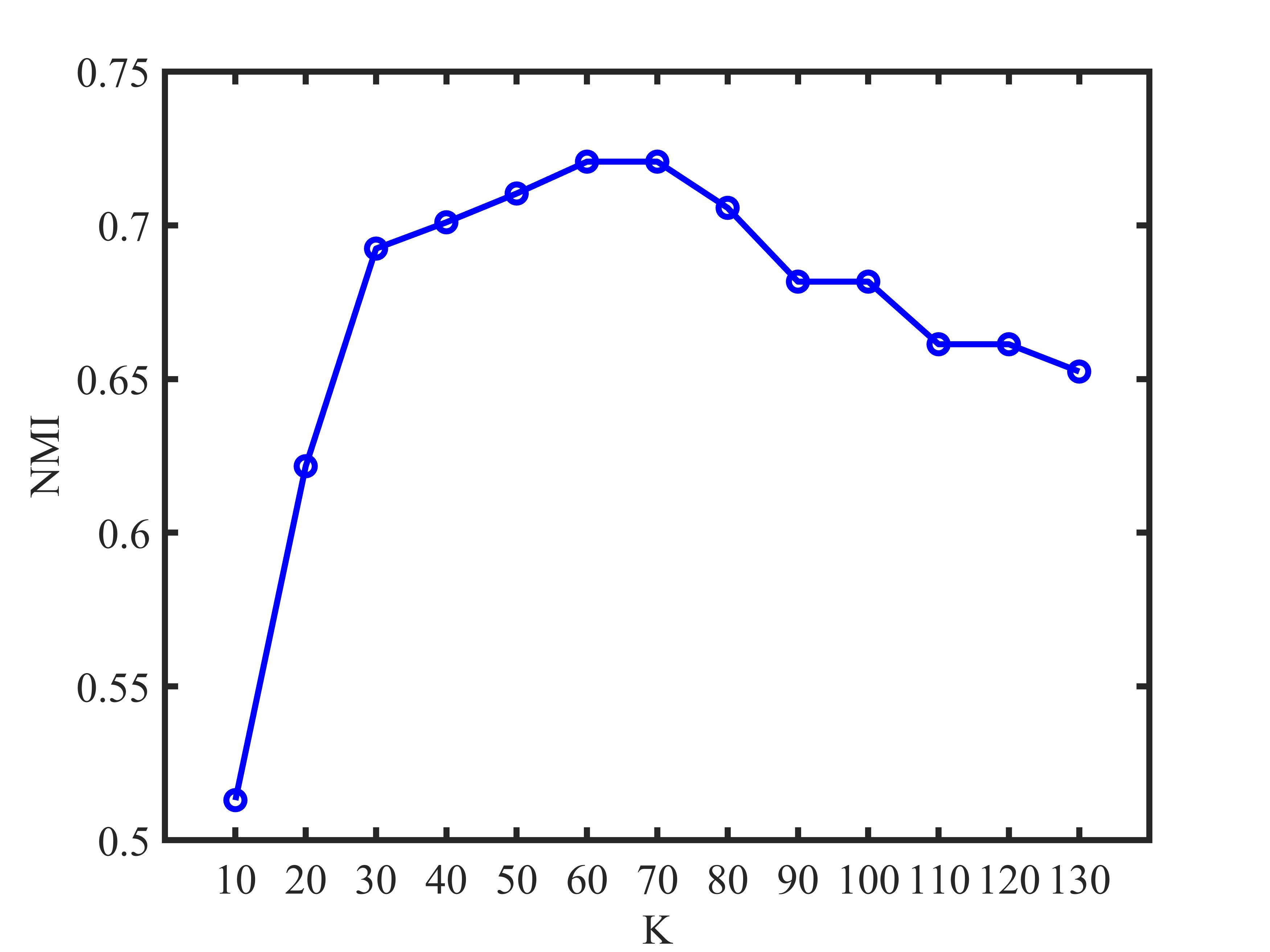}
}
\subfigure[PUR in 3Source dataset]{
\centering
\includegraphics[width=0.31\textwidth]{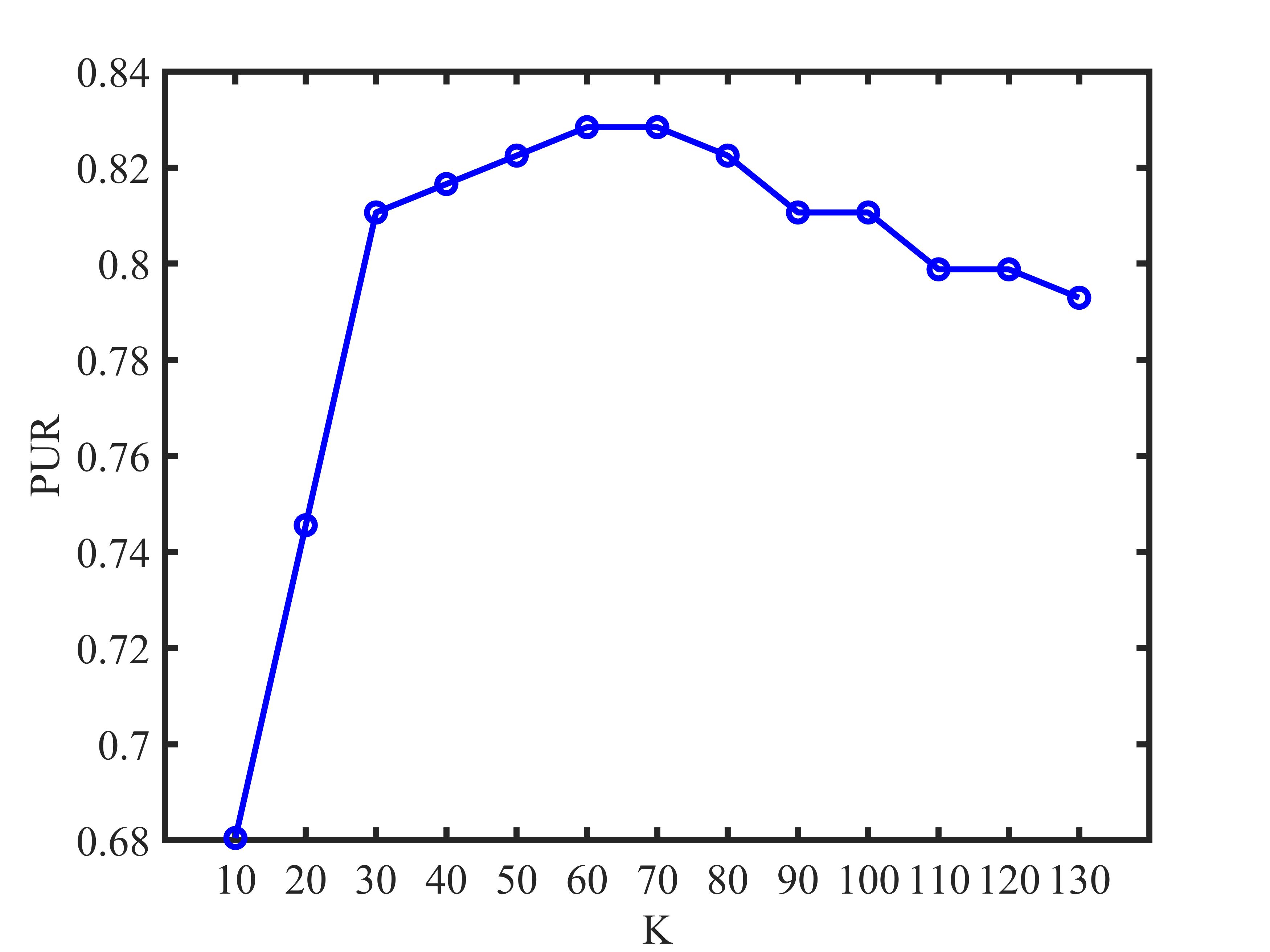}
}

\subfigure[ACC in Handwritten dataset]{
\centering
\includegraphics[width=0.31\textwidth]{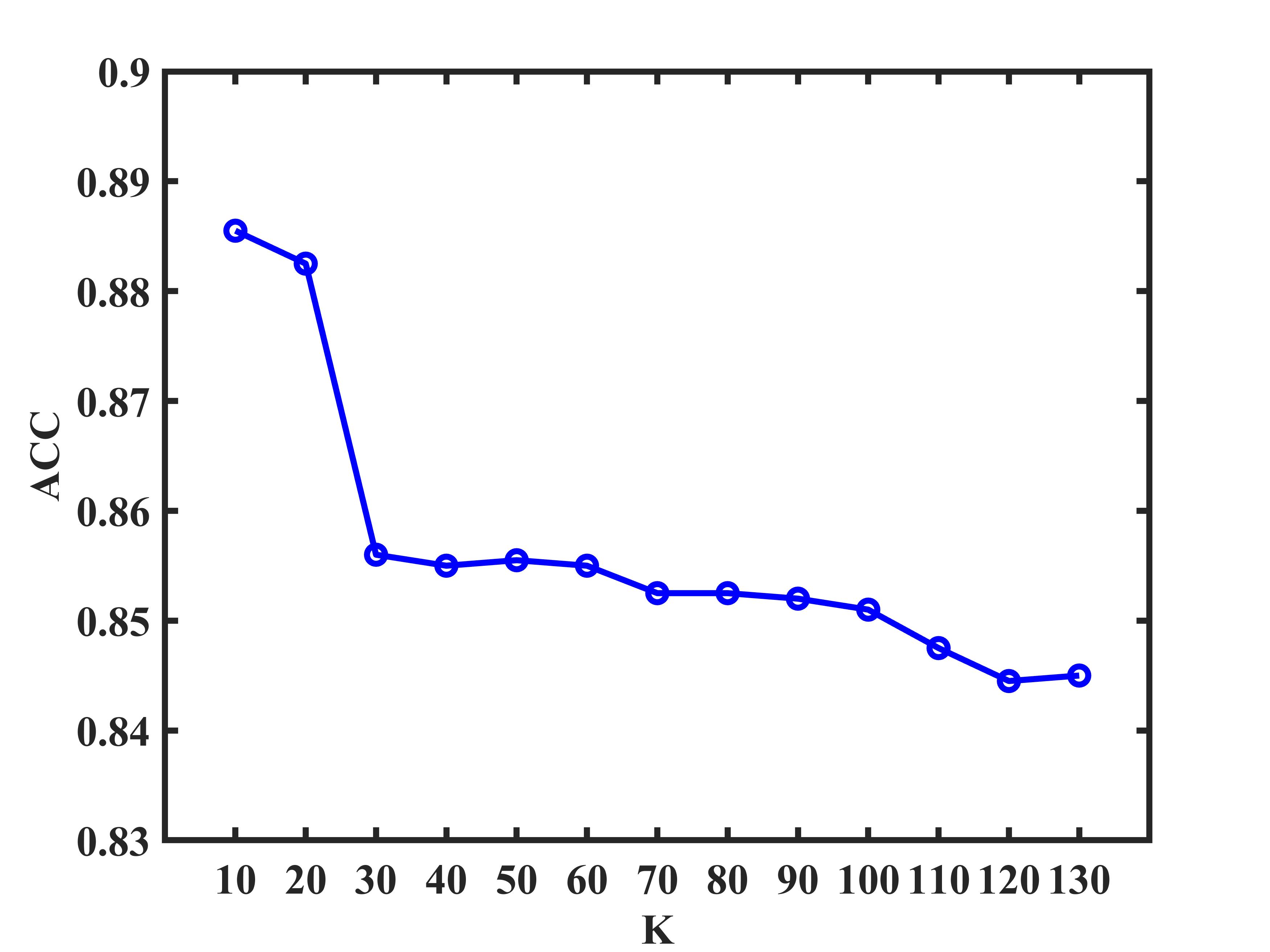}
}
\subfigure[NMI in Handwritten dataset]{
\centering
\includegraphics[width=0.31\textwidth]{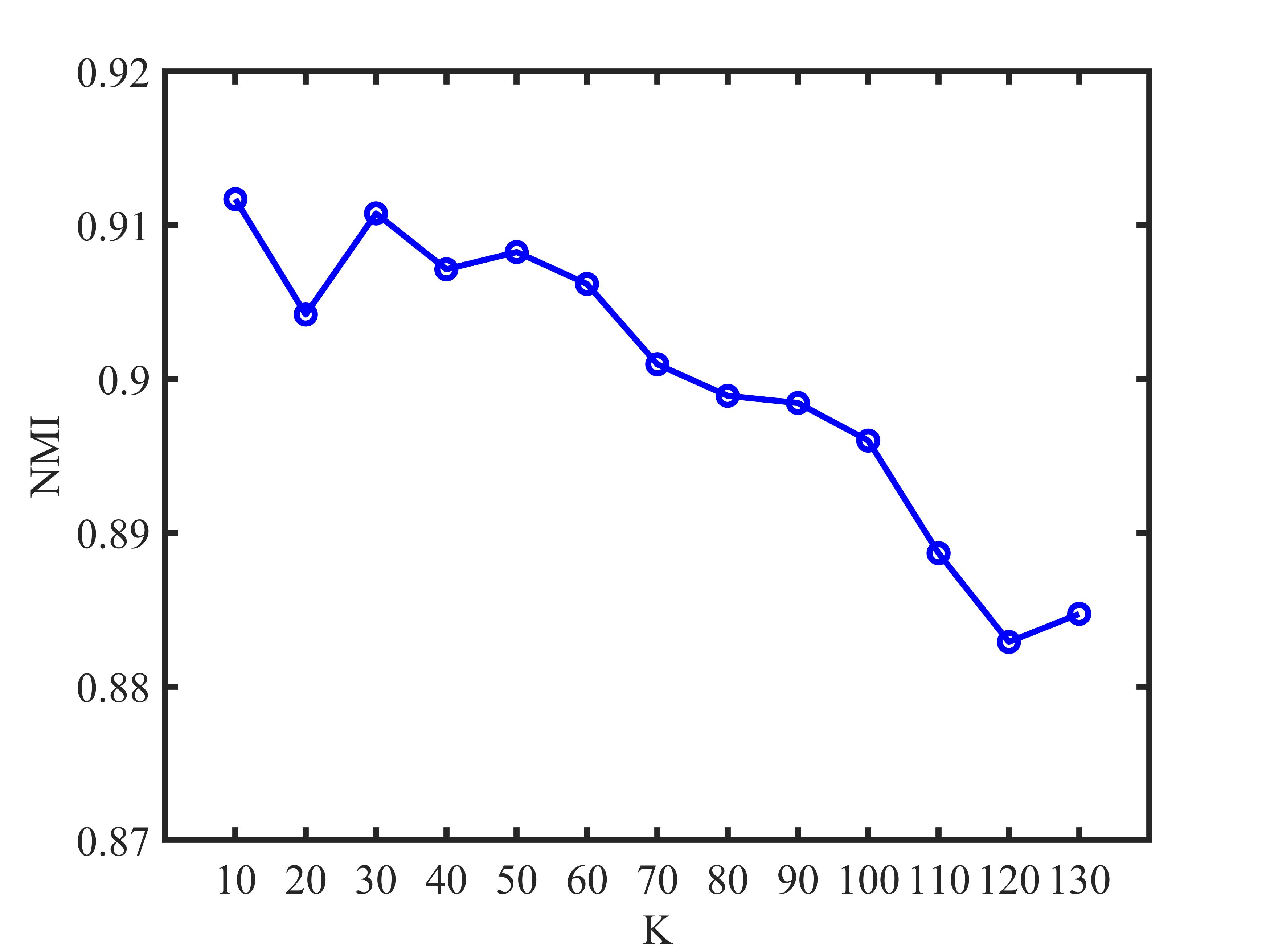}
}
\subfigure[PUR in Handwritten dataset]{
\centering
\includegraphics[width=0.31\textwidth]{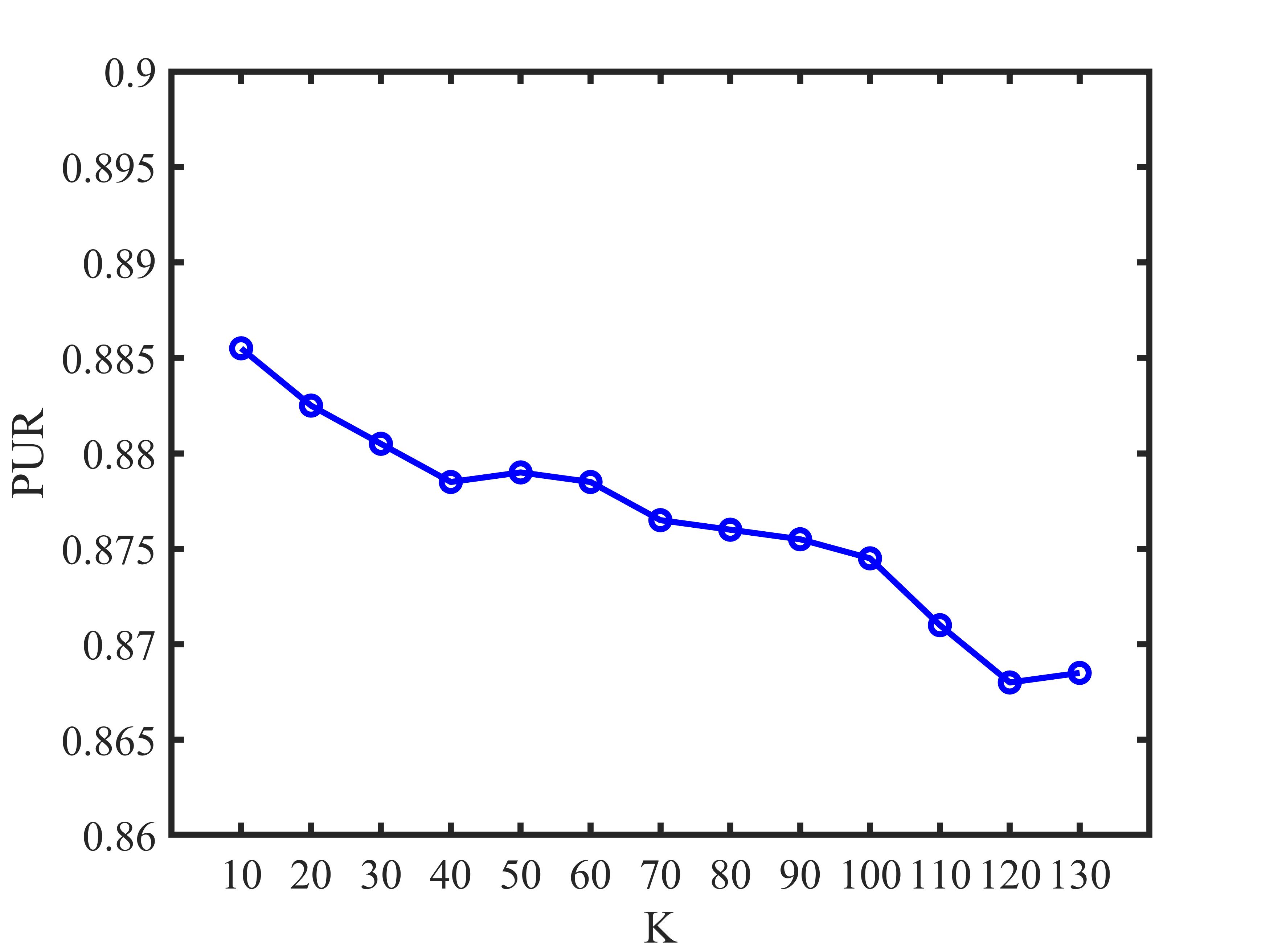}
}

\subfigure[ACC in NGs dataset]{
\centering
\includegraphics[width=0.31\textwidth]{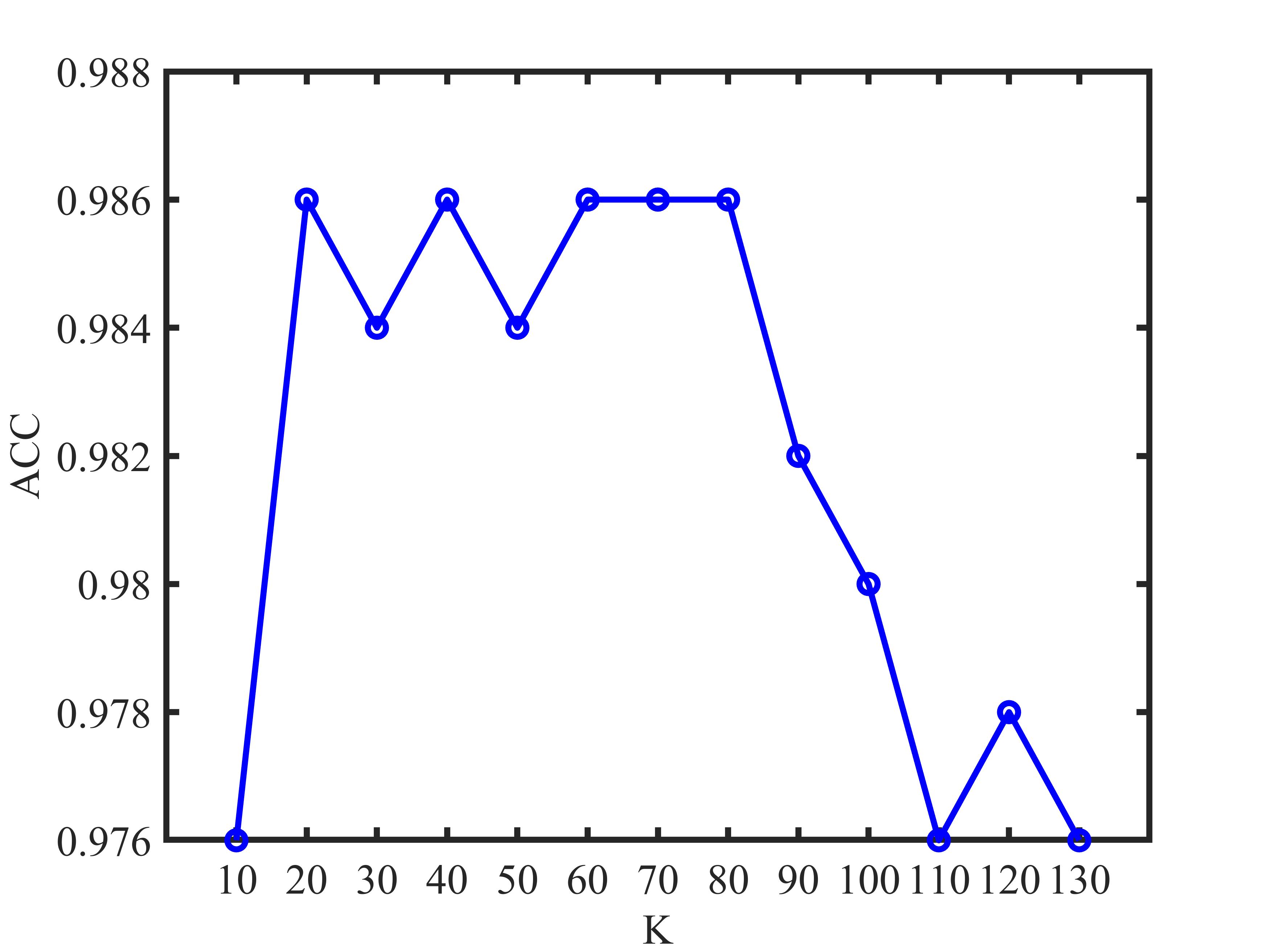}
}
\subfigure[NMI in NGs dataset]{
\centering
\includegraphics[width=0.31\textwidth]{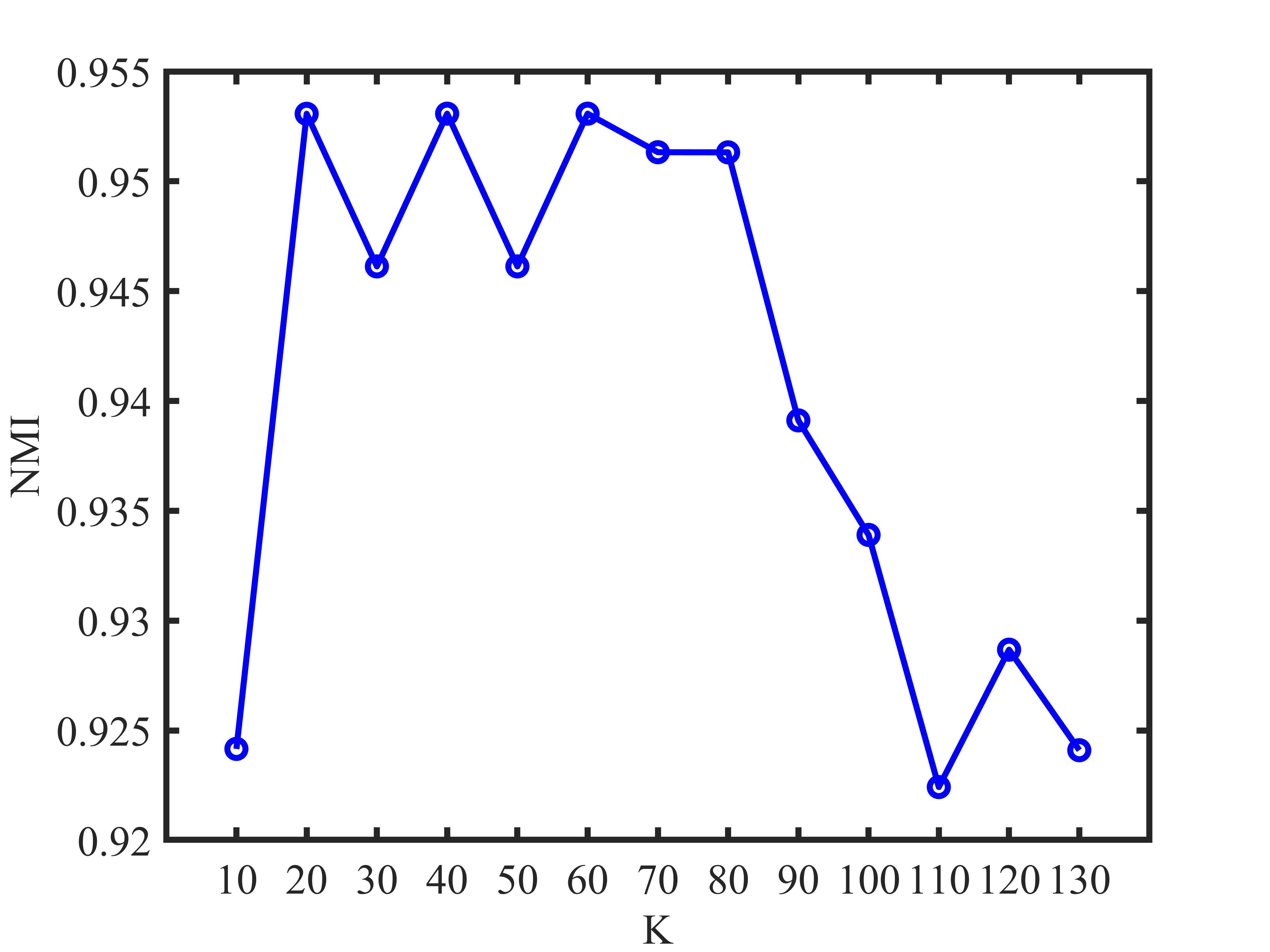}
}
\subfigure[PUR in NGs dataset]{
\centering
\includegraphics[width=0.31\textwidth]{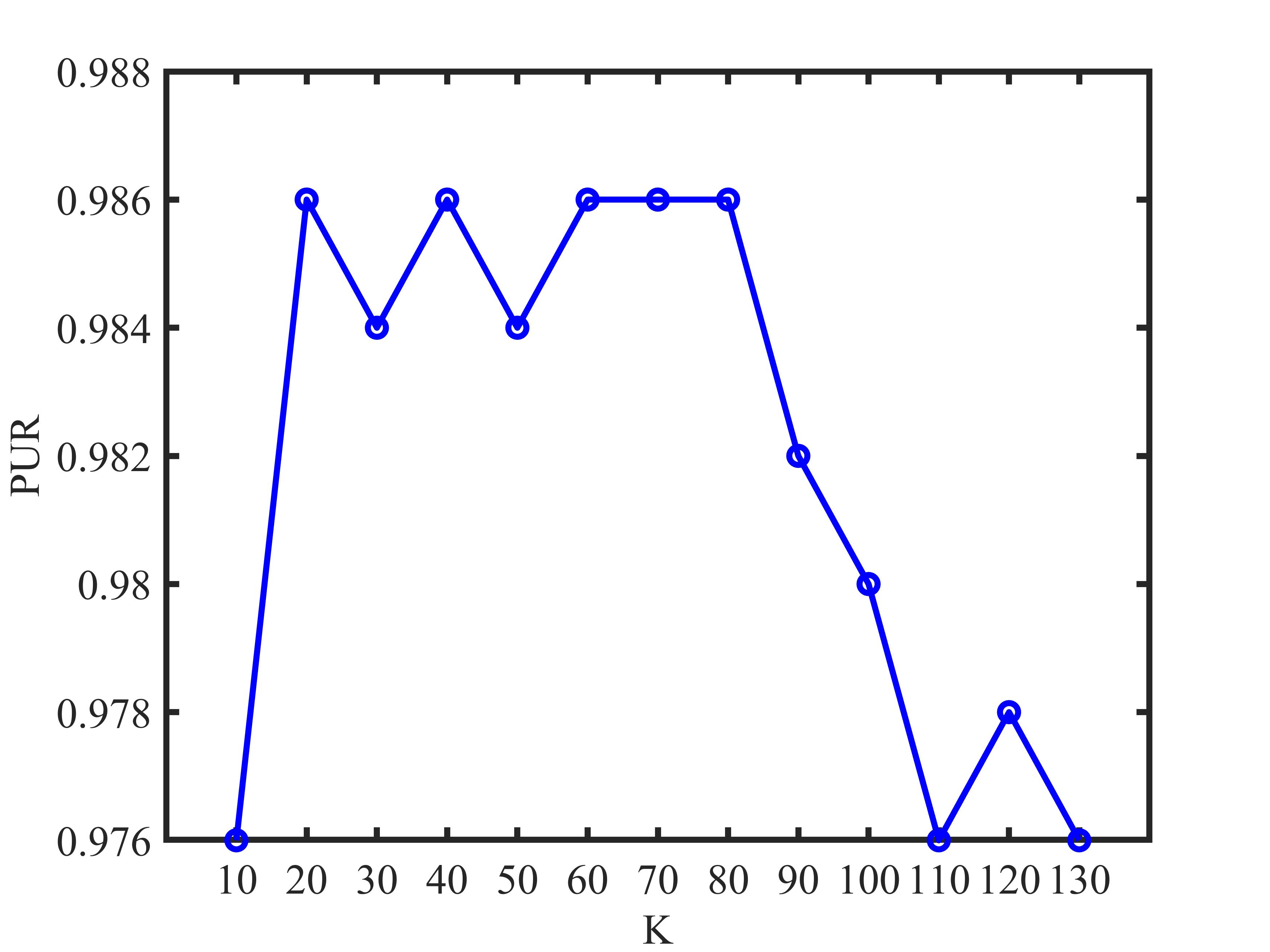}
}

\caption{K parameter analysis of LRC-MCF on four datasets}
\label{parameter}
\end{figure*}

The proposed LRC-MCF is based on the locality relationships constraint, thus we need to mainly discuss the sensitivity analysis on the number $K$ of nearest neighbors used in LRC-MCF in this section. Specifically, we conduct the clustering experiments for LRC-MCF on BBC, 3Source, HW2sources and NGs datasets, then the average value of all evaluation indexes is used as final criteria. Fig. \ref{parameter} plots the clustering results in terms of ACC, NMI, and PUR on these four datasets, under different $K$ in \{10, 20, 30, 40, 50, 60, 70, 80, 90, 100, 110, 120, 130\}. Through these variation trends of ACC, NMI, and PUR over these four datasets, it's easy to find that our method is relatively smooth to the $K$ over the relatively large ranges of values, which indicates that the performance is not so sensitive to those hyper-parameters. More importantly, there exists a wide range for each hyper-parameter in which relatively stable and good results can be readily obtained.

\subsection{Convergence Analysis}

\begin{figure*}[htbp]
\centering
\subfigure[BBCSport dataset]{
\centering
\label{fig-1}
\includegraphics[width=0.45\textwidth]{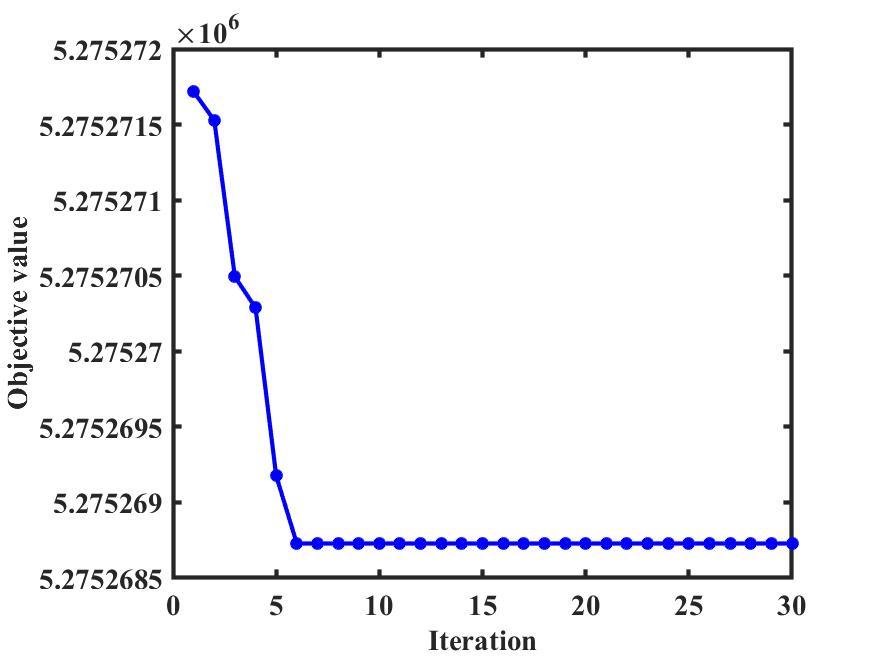}
}
\centering
\subfigure[Cora dataset]{
\centering
\includegraphics[width=0.45\textwidth]{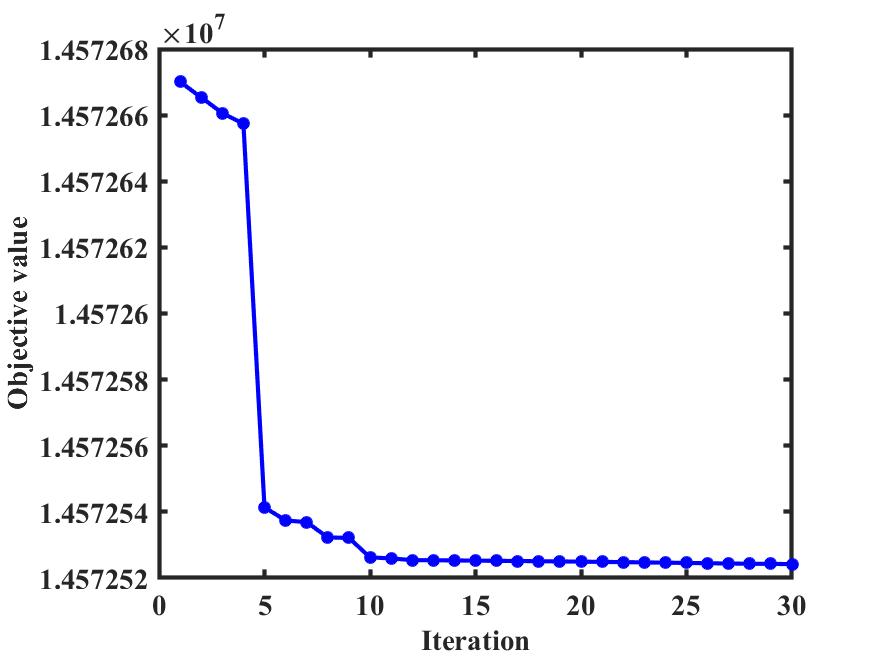}
}
\subfigure[Handwritten dataset]{
\centering
\includegraphics[width=0.45\textwidth]{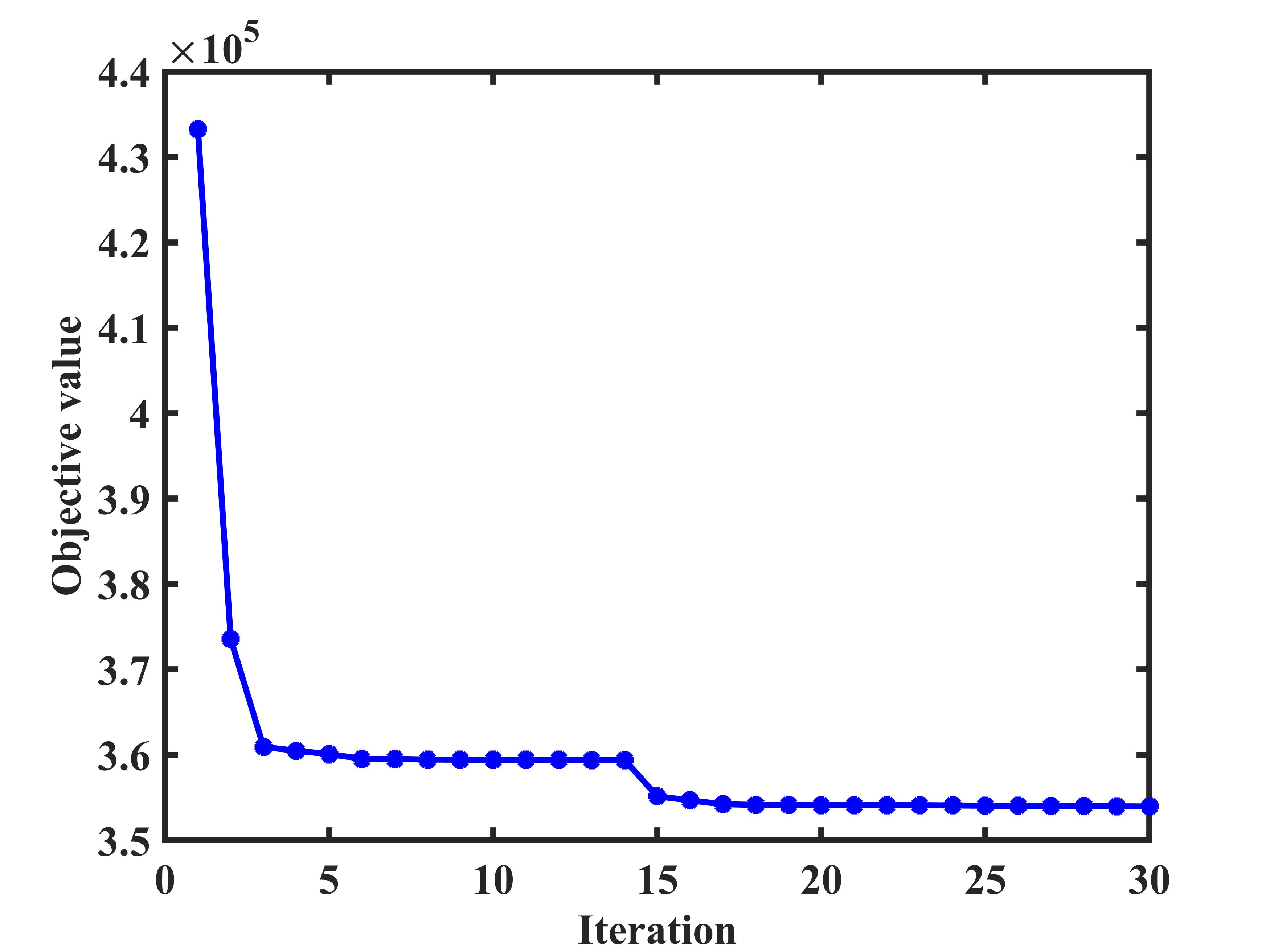}
}
\subfigure[NGs dataset]{
\centering
\label{fig-4}
\includegraphics[width=0.45\textwidth]{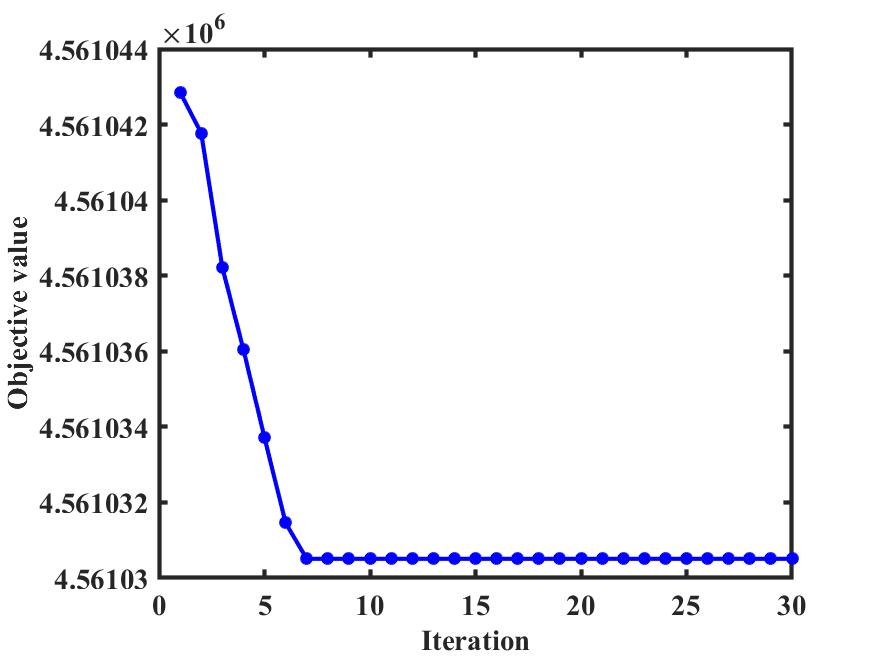}
}

\caption{Objective values of LRC-MCF on four datasets}
\label{convergence}
\end{figure*}

In this section, an iteration updating scheme is adopted by our proposed framework to obtain the optimal solution, thus convergence analysis is conducted to verify the convergence property of LRC-MCF in detail. We conduct various experiments on four datasets including BBCSport, Cora, Handwritten and NGs. Fig. \ref{convergence} plots the curves of objective values of LRC-MCF with the increase of iterations with responding to the above four datasets. According to the variation trend of objective values of LRC-MCF in these sub-figures of Fig.\ref{fig-1}-Fig.\ref{fig-4}, we can readily find that the objective value decreases rapidly during the iterations on all the four benchmark datasets. Moreover, it's not difficult to observe that the convergence can be reached within the 10 steps of iterations. That is to say, LRC-MCF can converge within a limited number of iterations. It also might imply that LRC-MCF can  effectually reduce the redundancy of the learned representations. However, when facing the complex and large-scale datasets, it's necessary to combine it with deep learning technology \cite{zhu2020deep,zhu2021graph} to construct the distance function. In the future, we consider how to extend superior deep models into our work, which can be optimized by the end-to-end manner.

\subsection{Discussion}
As the experiment results in Tables \ref{ACC} - \ref{PUR} on multi-view clustering tasks, we can clearly find that LRC-MCF outperforms other comparing methods in most situations. From the above evaluations, it's readily seen that the embeddings obtained by our method could be more effective and suitable for multi-view features. Besides, other multi-view methods outperform the other single-view methods in most situations, which could show multi-view learning is a valuable research field indeed. Compared to single view methods, taking the complementary information among different views into consideration could achieve more excellent performance, and the key is how to integrate the complementary information among different views while preserving its intrinsic characteristic in each view. According to Fig. \ref{parameter}, we find that the number $K$ of nearest neighbors used in our method is relatively smooth to the $K$ over the relatively large ranges of values. Note that the experimental results of our proposed LRC-MCF on seven datasets are without fine-tuning, and usage of fine-tuning might further improve its performance. For the discussion of convergence, according to Fig.\ref{convergence}, it could be empirically verified that the proposed LRC-MCF converges once enough iterations are completed. That is, LRC-MCF could converge within a limited number of iterations.

\section{Conclusion}
In this paper, we propose a novel multi-view learning method to explore the problem of multi-view clustering, called Locality Relationship Constrained Multi-view Clustering Framework (LRC-MCF). LRC-MCF attempts to capture the relationship information among samples under the constraint of locality structure and integrate the relationship from different views into one common similarity relationships matrix. Moreover, LRC-MCF provide an adaptive weight allocation mechanism to take sufficient consideration of the importance of different views in finding the common-view similarity relationships. To effectually reduce the redundancy of the learned representations, the low-rank constraint on the common similarity matrix is additionally considered. Correspondingly, this paper provides an algorithm based on the alternating direction minimization strategy to iteratively calculate all variables of LRC-MCF. Finally, we conduct extensive experiments on seven benchmark multi-view datasets to validate the performance of LRC-MCF, and the experimental results show that the proposed LRC-MCF outperforms its counterparts and achieves comparable performance. In the future, we will incorporate our proposed LRC-MCF and graph-based deep models \cite{zhu2021deep} to further explore multi-view information in those complex situations.

\section*{Acknowledgment}
The authors would like to thank the anonymous reviewers for their insightful comments and the suggestions to significantly improve the quality of this paper. The work was supported by the National Natural Science Foundation of PRChina(72001191), Henan Natural Science Foundation(202300410442), Henan Philosophy and Social Science Program(2020CZH009)is gratefully acknowledged. We also would like to thank the anonymous.

\bibliographystyle{IEEEtran}
\bibliography{IEEEexample}
\vspace{12pt}

\begin{IEEEbiography}[{\includegraphics[width=1in,height=1.25in,clip,keepaspectratio]{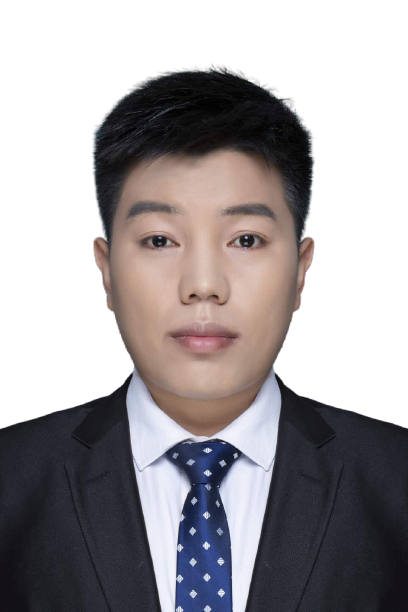}}]{Xiangzhu Meng} received his B.S. degree from Anhui University, in 2015, and the Ph.D. degree in Computer Science and Technology from Dalian University of Technology, in 2021. Now, he is a postdoctoral researcher with the Center for Research on Intelligent Perception and Computing, Institute of Automation, Chinese Academy of Sciences, China. He regularly publishes papers in prestigious journals, including Knowledge-Based Systems, Engineering Applications of Artificial Intelligence, Neurocomputing, etc. In addition, he serves as a reviewer for ACM Transactions on Multimedia Computing Communications and Applications. His research interests include multi-modal learning, deep learning, and vision-language modeling.
\end{IEEEbiography}

\begin{IEEEbiography}[{\includegraphics[width=1in,height=1.25in,clip,keepaspectratio]{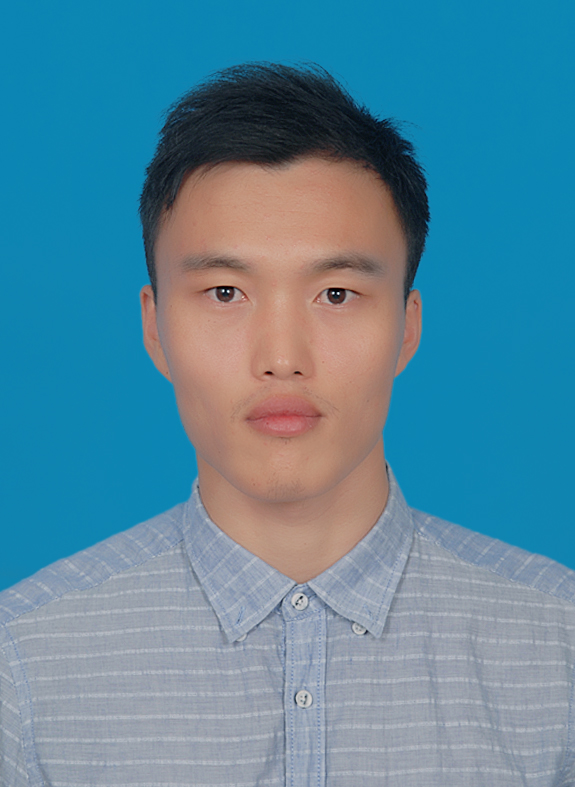}}]{Wei Wei} received the B.S. degree from the School of Mathematics and Information Science, Henan University, in 2012, and the Ph.D. degree in Institute of Systems Engineering from the Dalian University of Technology, in 2018. He is an associate professor of the Center for Energy, Environment \& Economy Research, College of Tourism Management, Zhengzhou University. His research interests include energy policy analysis, text mining, machine learning, and artificial intelligence.
\end{IEEEbiography}

\begin{IEEEbiography}[{\includegraphics[width=1in,height=1.25in,clip,keepaspectratio]{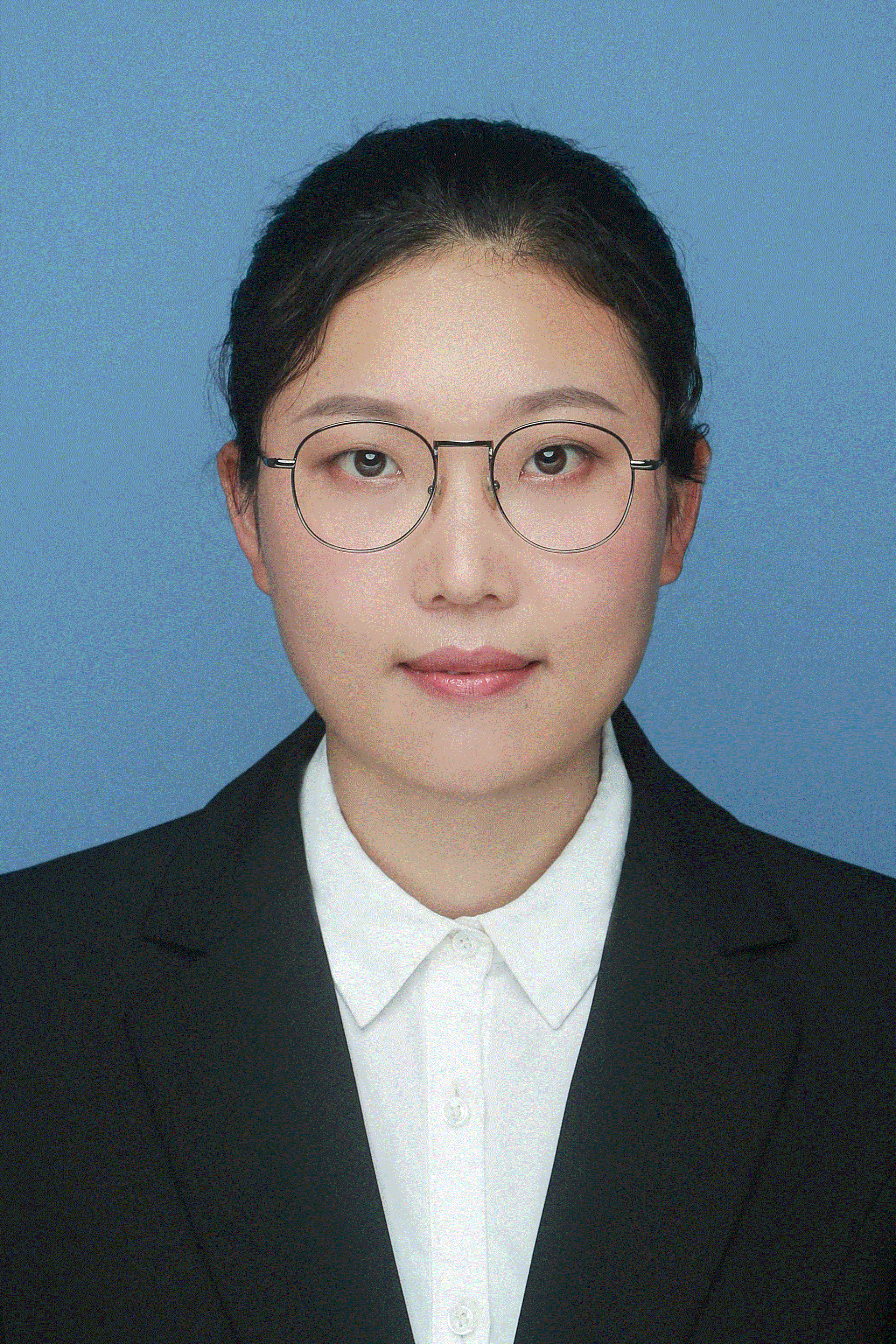}}]{Wenzhe Liu} received her BS degree from Qingdao University of Science and Technology, in 2013 and received MS degree from Liaoning Normal University . Now she is working towards the PHD degree in School of Computer Science and Technology, Dalian University of Technology, China. Her research interests include mulit-view learning, deep learning and data mining.
\end{IEEEbiography}

\end{document}